\documentclass[10pt,journal,compsoc]{IEEEtran}

\ifCLASSOPTIONcompsoc
  \usepackage[nocompress]{cite}
\else
  \usepackage{cite}
\fi

\usepackage{graphicx}
\usepackage{textcomp}
\usepackage{color}
\usepackage{amsmath}
\usepackage{url}
\usepackage{bm}
\usepackage{multirow}
\usepackage{algorithm}
\usepackage{makecell}
\usepackage{subfigure}
\usepackage{algpseudocode}
\usepackage{verbatim}

\newcommand{\RomanNumeralCaps}[1]{\MakeUppercase{\romannumeral #1}}

\def\XTH{{\textperthousand}}

\usepackage{amsmath,amsfonts,bm}

\def\vp{{\bm{p}}}
\def\vq{{\bm{q}}}

\def\vu{{\bm{u}}}
\def\vv{{\bm{v}}}

\def\vx{{\bm{x}}}

\def\mA{{\bm{A}}}

\def\mF{{\bm{F}}}

\def\mH{{\bm{H}}}
\def\mI{{\bm{I}}}
\def\mJ{{\bm{J}}}

\def\mL{{\bm{L}}}
\def\mM{{\bm{M}}}

\def\gR{{\mathcal{R}}}

\def\KRT{{\mathbb{M}}}
\def\UV{{\bm{B}}}

\def\0{\mathbf{0}}
\def\1{\mathbf{1}}

\def\Figref#1{Figure~\ref{#1}}
\def\Secref#1{Section~\ref{#1}}
\def\Algref#1{Alg.~\ref{#1}}
\def\Tabref#1{Table~\ref{#1}}
\def\Eqref#1{Eq.~\eqref{#1}}
\def\NAME{{SRT}}
\def\WS#1{{\omega_{\textrm{#1}}}}
\def\PERR{{$P\textrm{-}error$}}

\begin{document}

\title{Supervision by Registration and Triangulation\\for Landmark Detection}

\author{Xuanyi Dong,
        Yi Yang,
        Shih-En Wei,
        Xinshuo Weng,
        Yaser Sheikh,
        Shoou-I Yu
\IEEEcompsocitemizethanks{
\IEEEcompsocthanksitem Xuanyi Dong and Yi Yang are with Centre for Artificial Intelligence, University of Technology Sydney, NSW, Australia. (e-mail: xuanyi.dong@student.uts.edu.au; yi.yang@uts.edu.au)\protect
\IEEEcompsocthanksitem Shoou-I Yu, Shih-En Wei, and Yaser Sheikh are with Facebook Reality Labs, Pittsburgh, USA. (e-mail: shoou-i.yu@fb.com; shih-en.wei@fb.com; yaser.sheikh@fb.com)\protect
\IEEEcompsocthanksitem Xinshuo Weng is with the Robotics Institute, Carnegie Mellon University, Pittsburgh, USA. (e-mail: xinshuow@cs.cmu.edu)\protect
}
}

\markboth{IEEE TRANSACTIONS ON PATTERN ANALYSIS AND MACHINE INTELLIGENCE, 10.1109/TPAMI.2020.2983935}{}

\IEEEtitleabstractindextext{
\begin{abstract}
We present Supervision by Registration and Triangulation (SRT), an unsupervised approach that utilizes unlabeled multi-view video to improve the accuracy and precision of landmark detectors. Being able to utilize unlabeled data enables our detectors to learn from massive amounts of unlabeled data freely available and not be limited by the quality and quantity of manual human annotations. To utilize unlabeled data, there are two key observations:
(\RomanNumeralCaps{1}) the detections of the same landmark in adjacent frames should be coherent with registration, i.e., optical flow.
(\RomanNumeralCaps{2}) the detections of the same landmark in multiple synchronized and geometrically calibrated views should correspond to a single 3D point, i.e., multi-view consistency.
Registration and multi-view consistency are sources of supervision that do not require manual labeling, thus it can be leveraged to augment existing training data during detector training. End-to-end training is made possible by differentiable registration and 3D triangulation modules. Experiments with 11 datasets and a newly proposed metric to measure precision demonstrate accuracy and precision improvements in landmark detection on both images and video.
\end{abstract}

\begin{IEEEkeywords}
Landmark Detection, Optical Flow, Triangulation, Deep Learning
\end{IEEEkeywords}
}

\maketitle

\IEEEdisplaynontitleabstractindextext
\IEEEpeerreviewmaketitle

\IEEEraisesectionheading{\section{Introduction}\label{sec:introduction}}

\IEEEPARstart{A}{ccurate} and precise landmark detection is an important component to high quality performance of many computer vision and computer graphics tasks, such as face tracking~\cite{cao20133d, wu2018deep, yoon2019self}, face reenactment~\cite{thies2016face2face}, and body tracking~\cite{joo2018total,alp2018densepose}.
In many of these tasks, the landmarks are used to either provide good initialization for subsequent processing~\cite{koestinger2011annotated,cao20133d,joo2018total,wu2018deep}, 
or as loss terms which anchor the tracked face/body in an energy minimization or deep learning scenario~\cite{thies2016face2face,alp2018densepose,yoon2019self}.
Therefore, any errors such as inaccurate landmarks or unstable landmarks could propagate through the entire pipeline and have adverse effects on the final results.
For example, in face or body mesh tracking, 2D landmarks are used as anchors to deform a template mesh to match the current observations, so inaccuracies and instabilities in landmark detections could propagate to the tracked mesh and generate perceptually jarring results~\cite{dong2018sbr}.

While significant amount of work has been done on image-based landmark detection~\cite{lv2017deep,sagonas2013300,xiong2013supervised,wei2016convolutional,newell2016stacked,fang2017rmpe}, 
landmark detection accuracy and stability is far from optimal.
In the ideal case, independent frame-by-frame detections on a video sequence should be as accurate as if a marker was physically attached to the face or body.
However, high-frequency jitter is still observed when visualizing the frame-by-frame detections from state-of-the-art models~\cite{dong2018sbr}, suggesting that detector performance can be further improved.

\begin{figure}[t!]
\center
\includegraphics[width=\linewidth]{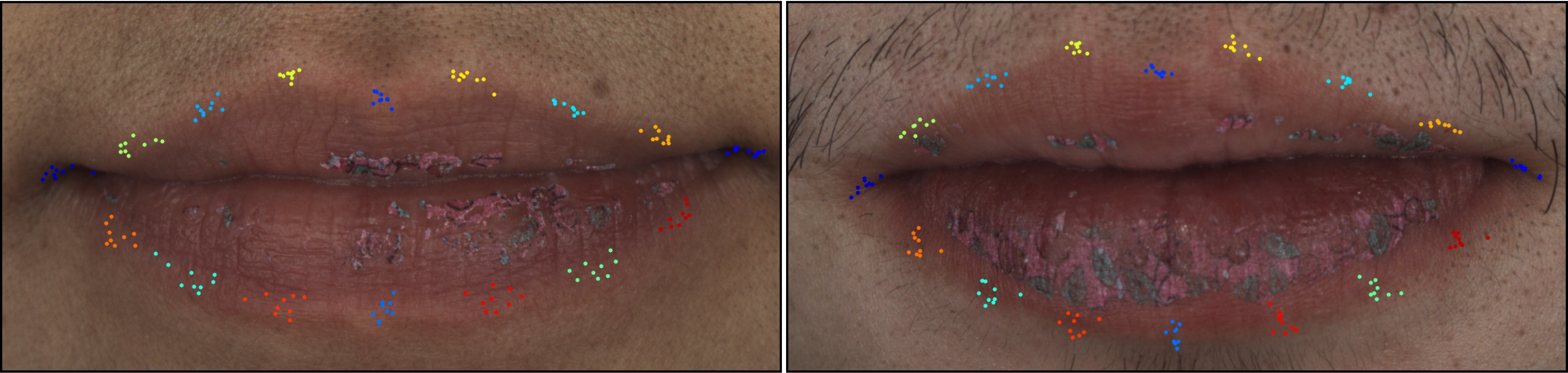}
\caption{
\textbf{Annotations are inconsistent.}
We show annotations of nine annotators on two images of the mouth. Each color indicates a different landmark.
Note the inconsistencies of annotations exist even on the more discriminative landmarks such as the corner of the mouth.
}
\label{fig:annotation-error}
\end{figure}

In order to improve detector performance, we first separate detector performance in two aspects, accuracy and precision, to facilitate analysis. An \textit{accurate} detector predicts landmarks that are consistent to manual annotations. A \textit{precise} detector localizes a landmark at the exact same semantic location across different input images. Note that an accurate detector is usually also precise, but a precise detector does not necessarily mean it is accurate, i.e., the detections can be consistent across different input images but still far from human annotations. One extreme example is the detector can always predict $(0, 0)$, which is semantically very consistent (very precise) but also very inaccurate. On the other hand, the accuracy metric heavily depends on the quality of human annotations which have inherent limits on consistency as shown in \Figref{fig:annotation-error}, and noisy annotations in the test set could make the detector accuracy look worse than it actually is as shown in \cite{dong2018sbr}. This is not the case for precision, where our proposed approach (see Section~\ref{sec:exp-setting}) can compute precision without relying on annotations, thus it is no longer affected by inconsistencies in annotations. In sum, both metrics have its merits and limitations, so in this paper, we measure both the accuracy and precision of the detector to get the most comprehensive story.

Based on the aforementioned metrics, we analyze the causes of unstable predictions, which could be due to: (\RomanNumeralCaps{1}) insufficient training samples, and (\RomanNumeralCaps{2}) imprecise annotations.
In \Figref{fig:Sync-Effect}, we analyzed the effect of varying the number of training samples and error of annotations on a synthetic dataset, which can provide ground-truth with zero annotation error. 
Results show that both accuracy and precision increases as we increase the number of training samples,
and as we add noise to the annotations used for training, both accuracy and precision drop significantly.
We observe that a model which was trained on annotation data with zero error achieves better performance than a model trained on twice the amount of noisy annotations.
This suggests that achieving high quality landmark detection might require very large amounts of imprecise human annotations, which is very labor intensive to collect.

\begin{figure*}[t]
\center
\includegraphics[width=\linewidth]{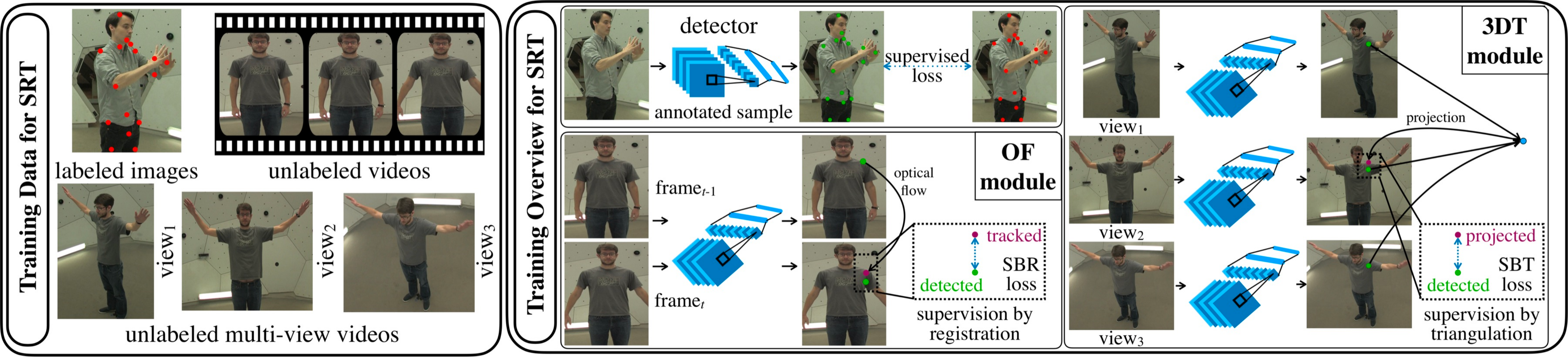}
\caption{
The \textbf{{Supervision by Registration and Triangulation} (\NAME) framework} takes labeled images and unlabeled synchronized and geometrically calibrated multi-view video as input to train an image-based landmark detector which is more precise on images/video, more stable on video, and also more consistent in multi-view scenarios. OF and 3DT stands for Optical Flow and 3D Triangulation respectively.
}
\label{fig:high-level}
\end{figure*}

Instead of completely relying on human annotations and being limited by their quality and quantity, we present {\it Supervision by Registration and Triangulation} (\NAME),
a method which augments the training loss function with supervision automatically extracted from {\it unlabeled} multi-view videos.
SRT consists of Supervision-by-Registration (SBR) and Supervision-by-Triangulation (SBT), which we detail in the following paragraphs.

For SBR, the consistency of (\RomanNumeralCaps{1}) the detections of the same landmark in adjacent video frames and (\RomanNumeralCaps{2}) registration, i.e., optical flow, is a source of supervision.
More specifically, a detected landmark at frame$_{t-1}$ followed by optical flow tracking between frame$_{t-1}$ and frame$_{t}$ should coincide with the location of the detection at frame$_{t}$.
Therefore, if the detections are incoherent with the optical flow, the mismatch is a supervisory signal enforcing the detector to be temporally consistent across frames, thus enabling a SBR-trained detector to better locate the correct location of a landmark which may be hard to annotate precisely.

For SBT, the consistency of (\RomanNumeralCaps{1}) the detections of the same landmark in different synchronized plus geometrically calibrated views and (\RomanNumeralCaps{2}) the 3D triangulation constraint, is a source of supervision.
More specifically, a detected landmark in different views should coincide with the reprojected landmarks calculated via 3D triangulation.
Therefore, if the detections are incoherent with the reprojected landmarks, the mismatch is a supervisory signal enforcing the detector to be spatially consistent across views.

The overview of our method is shown in \Figref{fig:high-level} and \Figref{fig:framework}.
Our end-to-end trainable model consists of three modules: a generic detector built on convolutional neural networks~\cite{newell2016stacked,wei2016convolutional}, a differentiable optical flow (OF) module, and a differentiable 3D triangulation (3DT) module.
During the forward pass, the OF module takes the landmark detections from the past frame and estimates their locations in the current frame.
The tracked landmarks are then compared with the detections on the current frame. The registration loss is defined as the offset between them.
The 3DT module takes the landmark detections from different views and estimates the 3D location.
The reprojected landmarks from the 3D location are then compared with the detections in each view. The multi-view loss is defined as the discrepancy between them.
In the backward pass, the gradients from the registration loss and the multi-view loss are back-propagated through the OF and 3DT modules to encourage spatial-temporal coherency in the detector.
The final output of our method is an enhanced image-based landmark detector, which has leveraged large amounts of unlabeled synchronized and geometrically calibrated multi-view video to achieve higher accuracy and precision in both images and videos, more stable predictions in videos, and more consistent predictions in different views.

Note that our approach is fundamentally different from post-processing such as temporal filtering, which often sacrifices precision for stability.
Our method directly incorporates the supervision of registration and multi-view coherency during model {\it training}, thus producing detectors that are inherently more stable.
Therefore, during testing time, neither post-processing, optical flow tracking, nor recurrent units are required.
Also note that SRT is not regularization, which limits the freedom of model parameters to prevent over-fitting.
Instead, SRT brings more supervisory signals from registration and triangulation to enhance the accuracy and precision of the detector.

In order to evaluate our claims, we perform a series of experiments to validate our method.
The questions we try to answer are as follows:

\noindent{(\RomanNumeralCaps{1})} Do both accuracy and precision improve through SRT?

\noindent{(\RomanNumeralCaps{2})} Which one is more useful, SBR or SBT? And do they complement each other?

\noindent{(\RomanNumeralCaps{3})} How do the quality and quantity of unlabeled videos affect the accuracy and precision of the detector?

\noindent{(\RomanNumeralCaps{4})} When do SBR and SBT not work?

Experiments on both regression and heatmap-based detectors
show that SBR and SBT are both helpful and can complement each other in improving accuracy and precision for landmark detection.
The improvement is most significant when the distribution of the unlabeled data is similar to the labeled data,
and inversely, when the distributions differ greatly, SBR and SBT may actually harm performance.
Also, there is a low chance (0.46\% in our optical flow experiments in \Secref{sec:exp-ablation}) that incorrect detections are still consistent with optical flow or multi-view constraints,
which could lead to the detector learning from incorrect supervision and hurting performance.

In sum, SRT has the following benefits:

\noindent{(\RomanNumeralCaps{1})} SRT can enhance the accuracy and precision of a generic landmark detector on both images and multi-view video without requiring additional labeled data.

\noindent{(\RomanNumeralCaps{2})} Since the supervisory signal of SRT does not come from annotations, SRT can leverage a very large amount of unlabeled synchronized and geometrically calibrated multi-view video to enhance the detector,
thus SRT is no longer limited by the quality and quantity of manual human annotations.

\noindent{(\RomanNumeralCaps{3})} SRT can be trained end-to-end.

\noindent{(\RomanNumeralCaps{4})} SRT is only applied during \emph{training} time, thus the prediction speed at test time is not affected.

\section{Related Work}\label{sec:relate-work}

Landmark detection algorithms are mainly applied to three modalities: images, video, and multi-view.
In images, the detector can only rely on the static image to detect landmarks, whereas in video the detector has additional temporal information to utilize.
In multi-view systems, additional geometric information is available to further boost the performance of a detector.
We will briefly compare our method with other detection algorithms in these three modalities.

\subsection{Image-based Landmark Detection}\label{sec:relate-work-image}

Despite some early works that learn to regress the coordinates of each landmark from hand-crafted features~\cite{xiong2013supervised}, recent landmark detection methods take advantage of end-to-end training from the deep convolutional neural network (CNN) model~\cite{wei2016convolutional,newell2016stacked}.
A typical method was to append an linear regression layer to predict the coordinates right after the CNN features and train the network in an end-to-end fashion.
To improve performance, one option is to cascade multiple CNN models to progressively refine the predictions~\cite{zhu2016unconstrained,dollar2010cascaded,dong2018san}.
For example, Yu~et~al.~\cite{yu2016deep} proposed a deformation network to incorporate geometric constraints in the CNN framework.
Zhu~et~al.~\cite{zhu2016unconstrained} leveraged cascaded regressors to handle extreme head poses and rich shape deformation.
Lv~et~al.~\cite{lv2017deep} presented a two-stage architecture to explicitly deal with the poor initialization problem.

Another category of landmark detection methods learn to predict a heatmap for each landmark~\cite{li2016face,wei2016convolutional,bulat2017far,nie2019hierarchical,wu2018look}.
Specifically, these methods define a Gaussian map at each ground-truth landmark coordinate, and the network aims to output that Gaussian map.
Wei~et~al.~\cite{wei2016convolutional} and Newell~et~al.~\cite{newell2016stacked} took the location with the highest response of the heatmap as the coordinate of the corresponding landmarks.
Li~et~al.~\cite{li2016face} enhanced the landmark detection through multi-task learning.
Bulat~et~al.~\cite{bulat2017far} proposed a robust network structure to utilize the advanced residual hourglass architectures.

Our proposed SRT is orthogonal to these image-based algorithms in that SRT can enhance both regression and heatmap-based detectors as we demonstrate in Section~\ref{sec:experiments}.

\subsection{Video-based Landmark Detection}\label{sec:relate-work-video}
Though image-based landmark detectors can achieve very good performance on images, sequentially running these detectors on each frame of multi-view videos in a tracking-by-detection fashion usually leads to jittering, unstable, and inconsistent detections.
One way to decrease jittering is to just initialize the tracker with the detection once then perform temporal tracking~\cite{baker2004lucas}, but this suffers from tracker drift. Once the tracker has failed in the current frame, it is difficult to make the correct prediction in the subsequent frames.
Therefore, hybrid methods~\cite{khan2017synergy,liu2017two,peng2016recurrent,girdhar2018detect} jointly utilize tracking-by-detection and temporal information in a single framework to predict more stable landmarks.
Peng~et~al.~\cite{peng2016recurrent} and Liu~et~al.~\cite{liu2017two} utilized recurrent neural networks to encode the temporal information across consecutive frames.
Khan~et~al.~\cite{khan2017synergy} utilized global variable consensus optimization to jointly optimize detection and tracking in consecutive frames.
Unfortunately, these methods require per-frame annotations, which are not only resource-intensive to acquire,
but also difficult to annotate consistently across frames, even for temporally adjacent frames.
Even thought SRT shares the same high-level idea of these algorithms by leveraging temporal coherency,
but SRT does not require any video-level annotation, and is therefore not limited by the availability and precision of human annotations.

Other approaches utilize temporal information in video to construct person-specific models~\cite{peng2018toward}.
Most of these methods usually leverage offline-trained static appearance models, i.e.,
the detector, which is used to generate initial landmark prediction, is not updated based on the tracking result in their algorithms, whereas SBR in our SRT dynamically refines the detector based on OF tracking results.
Self-training~\cite{zhu2007semi} can also be utilized for creating person-specific models, and was shown to be effective in pose estimation \cite{charles2016personalizing,shen2014unsupervised}.
However, unlike our method which can be trained end-to-end,~\cite{charles2016personalizing,shen2014unsupervised} did alternating bootstrapping to progressively improve the detectors. These methods make hard decisions on whether a pseudo-labeled sample should be added to the training set or not, and may also suffer from inaccurate gradient updates \cite{dong2018sbr}.

\subsection{Multi-view Landmark Detection}\label{sec:relate-work-view}

Leveraging geometrically calibrated and synchronized multi-view images enable us to utilize epipolar constraints to enhance the detector.
A straightforward solution is bootstrapping~\cite{simon2017hand}. Starting from an initial weak detector, triangulation and reprojection is used
to generate robust pseudo-labels on unlabeled images, which is then used to enhance the initial detector.
Other researchers took this approach one step further by incorporating the multi-view loss into their objective and train the model in an end-to-end fashion~\cite{suwajanakorn2018discovery,jafarian2018monet,zhang2018multiview}.
Suwajanakorn~et~al.~\cite{suwajanakorn2018discovery} applied a multi-view consistency loss between every view pair in their framework.
The authors of~\cite{jafarian2018monet,zhang2018multiview} leveraged the epipolar constraint to reduce the mismatch between predictions in two different views.
Rhodin~et~al.~\cite{rhodin2018learning} trained the system to predict the same pose in all views.
Amberg~et~al.~\cite{amberg2007reconstructing} proposed to align the raw pixels of corresponding points in different views for 3D face reconstruction.
Pavllo~et~al.~\cite{pavllo20193d} proposed a semi-supervised training strategy by minimizing the bone length via a 3D model.
SRT shares the same high-level idea of these methods by utilizing multi-view coherency,
and one advantage of SRT is we are able to utilize multiple views, instead of just two views (\cite{jafarian2018monet,zhang2018multiview}), jointly in one training iteration, thus being able to leverage information from multiple views at once.

\section{Methodology}

\begin{figure}[t]
\center
\includegraphics[width=\columnwidth]{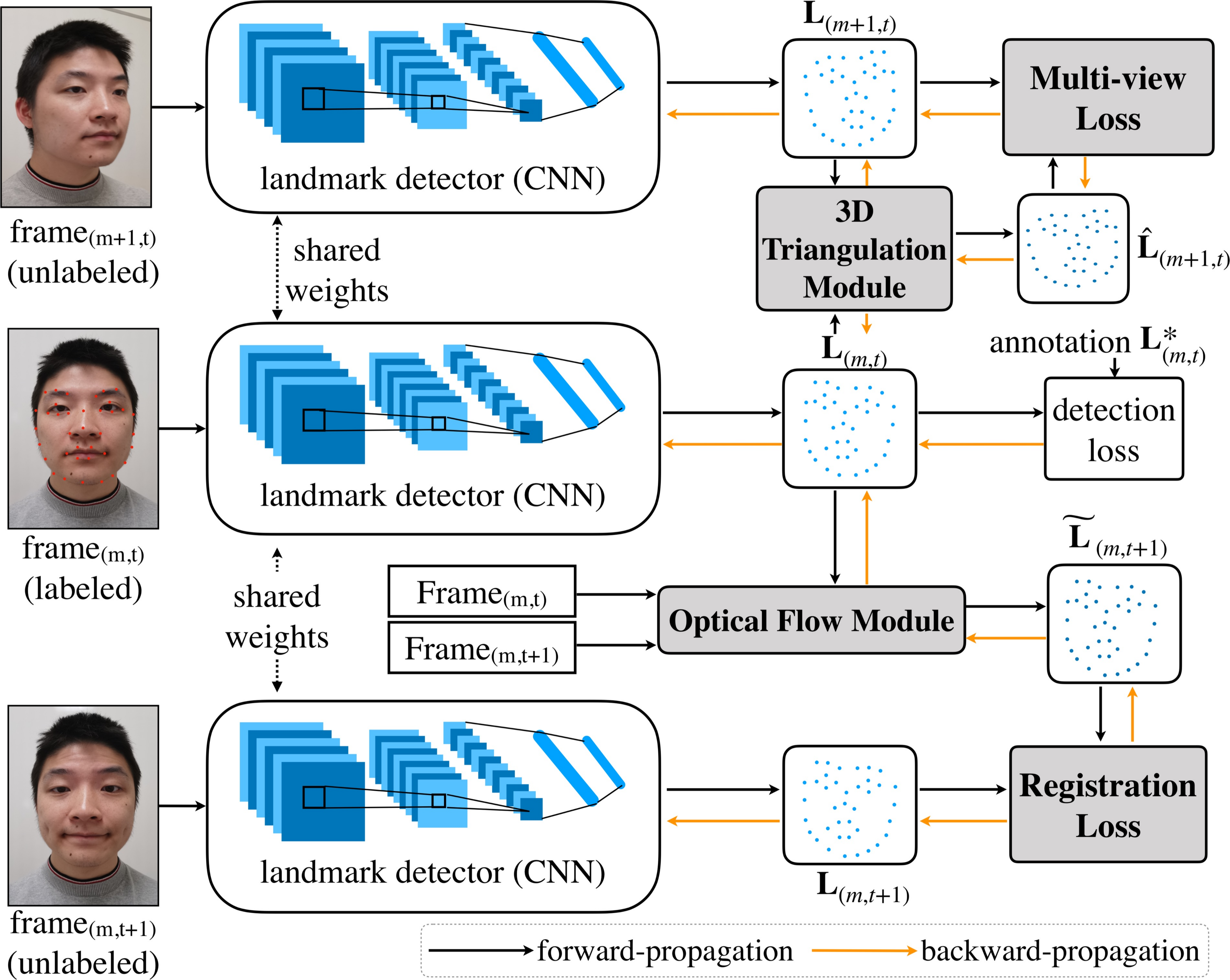}
\caption{
The \textbf{training} of {\NAME} with three complementary losses.
The key idea is that the supervision from registration and triangulation can directly back-propagate through the optical flow (OF) and 3D triangulation (3DT) modules respectively, thus enabling the detector before the OF and 3DT modules to receive gradients which encourage temporal and multi-view coherency across frames.
}
\label{fig:framework}
\end{figure}

{\NAME} consists of three complementary parts, the general landmark detector, the optical flow (OF) module, and the 3D triangulation (3DT) module, as shown in \Figref{fig:framework}.
Each part has a corresponding loss function.
The detection loss utilizes appearance from a single image and label information to learn a better landmark detector.
The registration and multi-view losses leverage the differentiable OF and 3DT modules respectively to enforce the predictions in neighboring frames and different views to be consistent.
The key components to {\NAME} are the OF and 3DT modules, which we describe in detail in the following sections.

\subsection{Differentiable Optical Flow Module}\label{sec:sbr-introduce}

For SBR, one essentially needs a differentiable OF module that takes as input a location on the image and outputs the OF at that location. We present two possible directions to implement this module.

The first direction is to decompose an existing OF algorithm
into a series of differentiable functions, thus enabling us to directly
incorporate the OF algorithm into deep network training.
In \Secref{sec:lkof}, we use the Lucas Kanade optical flow algorithm \cite{lucas1981iterative} as an example.
However, this method tends to be too computationally inefficient in practice.
Even though the theoretical computation of optical flow is negligible compared to the complexity of evaluating a CNN,
the actual speed of computing OF is significantly lower than evaluating a CNN.
This is mainly due to existing GPU libraries being highly optimized for common CNN modules such as convolutions, but not optimized for custom modules such as differentiable Lucas Kanade tracking.
Therefore, we also propose another direction which can efficiently approximate OF.

The second direction is to approximate OF at sub-pixel locations through bilinear interpolation on a pre-computed flow field, which is still a fully differentiable process.
This enables us to remove OF compute time out of network forward and backward passes, thus providing significant boosts in training speeds (please see analysis in supplementary material).
It also enables us to utilize more sophisticated and potentially non-differentiable OF algorithms.

We now detail each direction in the following sections and explain the key notations in \Tabref{table:notation}.

\subsubsection{Differentiable Lucas Kanade OF}\label{sec:lkof}
Motivated by~\cite{chang2017clkn}, we decompose the Lucas Kanade OF algorithm into a series of differentiable steps,
such that gradients can back-propagate through the optical flow module.

Given the feature $\mF_{(m,t-1)}$\footnote{The features can be RGB images or outputs of convolutional layers.} from frame$_{(m,t-1)}$ in view$_{m}$ (the $m$-th view) at time$_{t-1}$ {(the $t\textrm{-}1$-th timestamp)} and feature $\mF_{(m,t)}$ from frame$_{(m,t)}$, we estimate the motion for a small patch near $\vx_{(m,t-1)}=[x,y]^T$ from frame$_{(m,t-1)}$.
The motion model is represented by the displacement warp function $W(\vx;\vp)$.
A displacement warp contains two parameters $\vp=[p_{1}, p_{2}]^T$, and can be formulated as $W({\vx};{\vp})=[x + p_1,y + p_2]^{T}$.
We leverage the inverse compositional algorithm \cite{baker2004lucas} for our OF operation. It finds the motion parameter $\vp$ by minimizing
{
\begin{align}\label{eq:optical-flow}
\sum_{\vx\in\Omega} \alpha_{\vx}\parallel \mF_{(m,t-1)}(W(\vx;\Delta\vp)) - \mF_{(m,t)}(W(\vx;\vp))\parallel^{2} ,
\end{align}
}\noindent with respect to $\Delta\vp$. Here, $\Omega$ is a set of locations in a patch centered at $\vx_{(m,t-1)}$, and $\alpha_{\vx} = \exp(-\frac{||\vx-\vx_{(m,t-1)}||_2^2}{2\sigma^2})$ is the weight value for $\vx$ determined by the distance from $\vx_{(m,t-1)}$ to down-weight pixels further away from the center of the patch.
After obtaining $\Delta\vp$, we update $\vp$ as follows:
{
\begin{align}\label{eq:iter}
W(\vx;\vp) \leftarrow W(W(\vx;\Delta \vp)^{-1};\vp) = \left[ \begin{array}{ccc} x+p_1-\Delta p_1 \\ y+p_2-\Delta p_2 \\ \end{array} \right].
\end{align}
}\noindent ${\vp}$ is an initial motion parameter ($\vp$ = $[0,0]$ in our case), which will be iteratively updated by \Eqref{eq:iter} until convergence.

For each iteration, $\Delta\vp$ is computed through minimizing the first order Taylor expansion of \Eqref{eq:optical-flow}:
{
\begin{align}\label{eq:first}
\sum\nolimits_{\vx\in\Omega}~&\alpha_{\vx}\parallel \mF_{(m,t-1)}(W(\vx;\0))  \nonumber \\
                  + &\nabla\mF_{(m,t-1)}\frac{\partial W}{\partial\vp}\Delta\vp - \mF_{(m,t)}(W(\vx;\vp))\parallel^{2} .
\end{align}
}\noindent The $\Delta\vp$ which minimizes \Eqref{eq:first} is:
{
\begin{align}\label{eq:solution}
\Delta \vp = &\mH^{-1} \sum\nolimits_{\vx\in\Omega}J(\vx)^{T} \alpha_{\vx}   \nonumber \\
      \times &(\mF_{(m,t)}(W(\vx;\vp)) -  \mF_{(m,t-1)}(W(\vx;\0))) ,
\end{align}
}\noindent where $\mH$=$\mJ^{T} \mA \mJ$ $\in$ $\gR^{2\times2}$ is the Hessian matrix.
$\mJ \in \gR^{C|\Omega|\times2}$ is the vertical stacking of $J(\vx) \in \gR^{C\times2}$, $\vx \in \Omega$, which is the Jacobian matrix of $\mF_{(m,t-1)}(W(\vx;\0))$.
$C$ is the number of channels of $\mF$.
${\mA}$ is a diagonal matrix, where elements in the main diagonal are the $\alpha_{\vx}$'s corresponding to the $\vx$'s used to compute each row of $\mJ$. 
$\mH$ and $\mJ$ are constant over iterations and can thus be pre-computed.

\begin{algorithm}[t!]
\small
\caption{
The differentiable Lucas Kanade OF operation.
}
\label{alg:OF-operation}

  \begin{algorithmic}
    \Require $\mF_{(m,t-1)}$, $\mF_{(m,t)}$, and $\vp = [0,0]$
    \Require the $k$-th landmark location $\vx$, i.e., $\mL_{(m,t-1,k)}$ 
    \State 1. Extract template feature from $\mF_{(m,t-1)}$ centered at $\vx$
    \State 2. Calculate the gradient of the template feature
    \State 3. Compute the Jacobian and Hessian matrices, $\mJ$ and $\mH$
    \For{iter=1; iter $\leq$ max; iter++}
    	\State 4. Extract target feature from $\mF_{(m,t)}$ centered at $\vx+\vp$
        \State 5. Compute difference of template and target features
        \State 6. Compute $\Delta\vp$ using Eq.~\eqref{eq:solution}
        \State 7. Update the motion model $\vp$ using Eq.~\eqref{eq:iter}
    \EndFor
    \Ensure the $k$-th landmark location at frame$_{(m,t)}$ : $\vx + \vp$
  \end{algorithmic}
  
\end{algorithm}

We describe the detailed steps of the Lucas Kanade OF operation in \Algref{alg:OF-operation}.
We define the OF operation as $\widetilde{{\mL}}_{(m,t,k)}=G_{OF}(\mF_{(m,t-1)}, \mF_{(m,t)}, {\mL}_{(m,t-1,k)})$.
This function takes the detected location of the $k$-th landmark in frame$_{(m,t-1)}$: ${\mL}_{(m,t-1,k)} \in \gR^{2}$, as input computes the landmark location $\widetilde{{\mL}}_{(m,t,k)}$ for the next (future) frame.
Since, all steps in the OF operation are differentiable, the gradient can back-propagate to the facial landmark detections and the feature maps.

We add a very small value to the diagonal elements of $\mH$. This ensures that $\mH$ is invertible.
Also, in order to crop a patch at a sub-pixel location $\vx$, we use the spatial transformer network \cite{jaderberg2015spatial} to calculate the bilinear interpolated values of the feature maps.

\noindent\textbf{Loss calculation}.
SBR supervision comes from the discrepancy between the predicted coordinates ${\mL}_{(m,t,k)}$ and the estimated coordinates $\widetilde{{\mL}}_{(m,t,k)}$.
The loss function that calculates this discrepancy can vary for regression and heatmap-based detectors. We will introduce our choices in \Secref{sec:supervisionRT-introduction}.

\subsubsection{Bilinear Interpolation Approximation of OF}\label{sec:method-approx-of}

Another method to compute OF is to approximate OF at sub-pixel locations through bilinear interpolation on a pre-computed flow field, which is still a fully differentiable process.
An issue is that such approximation with bilinear interpolation might lead to drop in optical flow accuracy.
Fortunately, we empirically show that the OF computed from bilinear interpolation is very accurate (see details in \Secref{sec:exp-ablation}.)
Given that the OF error due to interpolation is minimal, this enables us to leverage many more sophisticated OF algorithms.

\subsection{Differentiable 3D Triangulation Module}\label{sec:sbt-introduction}

To incorporate multi-view consistency, we design a 3DT module through which we can perform back-propagation.
There are three steps: triangulation, reprojection, and loss calculation.
Given the predicted coordinate of a landmark from all synchronized and calibrated views, we first estimate the 3D coordinates through triangulation.
Then, the 3D point is reprojected back into each camera view.
Finally, the location difference between the reprojected and the detected points are used to compute the loss.
These steps are repeated for all landmarks. Details for each step are as follows.

\begin{table}[t!]
\setlength{\tabcolsep}{1pt}
\centering
\caption{
Explanation of notations in this manuscript.
}
\begin{tabular}{|c|c|} \hline\hline
      Notation                & Definition                   \\\hline
      $m$                     & the index for views           \\
      $t$                     & the index for time frames      \\
      $k$                     & the index for landmarks        \\
  $\mF_{(m,t)}$               & the feature tensor from the $t$-th frame in the $m$-th view \\
     $\vx$                    & a 2D coordinate              \\
     $\vp$                    & a 2D displacement            \\
    $W(\vx,\vp)$              & translate the coordinate $\vx$ with $\vp$ \\
    $\Delta \vp$              & a 2D vector to update $\vp$  \\
    $\mH$                     & the Hessian matrix           \\
    $K$                       & the number of landmarks      \\
  $\mL_{(m,t,k)}$             & the predicted $k$-th landmark at frame$_t$ in view$_m$    \\
$\widetilde{\mL}_{(m,t,k)}$   & the OF-tracked $k$-th landmark at frame$_t$ in view$_m$   \\
$\hat{\mL}_{(m,t,k)}$         & the 3DT-computed $k$-th landmark at frame$_t$ in view$_m$  \\
${\mL}^{*}_{(m,t,k)}$         & ground truth label of $k$-th landmark at frame$_t$ in view$_m$   \\
  $\mM_{(m,t,k)}$             & the $k$-th predicted heatmap at frame$_t$ in view$_m$  \\
$\KRT_{m} \in \gR^{3\times4}$ & the camera transformation matrix for the $m$-th view   \\
\hline
\end{tabular}
\label{table:notation}
\end{table}

\noindent\textbf{3D Triangulation}.
We solve for the 3D location of a landmark through Direct Linear Transformation~\cite{hartley2003multiple}.
Given the camera transformation matrices $\KRT_{m} \in \gR^{3\times4}$ for all $M$ views, and detections in each view $\mL_{(m,t,k)} \in \gR^{2}$, $1\le m \le M$, where $m$ indicates the index of views, we can calculate the 3D landmark $\mL_{(t,k)}^{{\textrm{3D}}} \in \gR^{3}$ as follows:
\begin{align}
    \vu_{m} = \KRT_{m}[0,:] - \KRT_{m}[2,:] \cdotp \mL_{(m,t,k)}[0] & \in \gR^{4}       ,\label{eq:3DT-U}\\
    \vv_{m} = \KRT_{m}[1,:] - \KRT_{m}[2,:] \cdotp \mL_{(m,t,k)}[1] & \in \gR^{4}       ,\label{eq:3DT-V}\\
    \UV = [\vu_{1} | \vu_{2} | ... | \vu_{m} | \vv_{1} | \vv_{2} |, ..., | \vv_{m}]^T & \in \gR^{2M\times4}   ,\label{eq:3DT-UV}\\
    \mL_{(t,k)}^{{\textrm{3D}}} = \left( \UV[:,{:}3]^T ~ \UV[:,{:}3] \right)^{-1} ~ \UV[:,{:}3&]^{T} ~ (-\UV[:,3])           .\label{eq:3DT-L}
\end{align}
\noindent
$\KRT_{m}[i,:]$ indicates the $i$-th row of $\KRT_{m}$.
$\UV[:,{:}i]$ indicates columns 0 to $i$-$1$ of $\UV$.
$\UV[:,i]$ indicates the $i$-th column of $\UV$. $\mL_{(m,t,k)}[i]$ denotes the $i$-th element of $\mL_{(m,t,k)}$.

\Eqref{eq:3DT-U} and \Eqref{eq:3DT-V} first compute the projection constraints for $\mL_{(t,k)}^{\textrm{3D}}$, i.e., $\vu_{m}[{:}3] \cdot \mL_{(t,k)}^{\textrm{3D}} + \vu_m[3] = 0$, where ``$\cdot$'' denotes the dot product and $\vu_{m}[{:}3]$ denotes the first three elements of $\vu_{m}$.
Then we stack all the constraints into $\UV \in \gR^{2M\times4}$ in \Eqref{eq:3DT-UV}.
Lastly, we solve for $\mL_{(t,k)}^{{\textrm{3D}}}$ with a least squares approach in \Eqref{eq:3DT-L}.

\noindent\textbf{3D Projection}.
The second step for SBT is 3D projection.
Given $\mL_{(t,k)}^{{\textrm{3D}}}$, 
We calculate the reprojected 2D landmark in view$_m$, $\hat{\mL}_{(m,t,k)}$, as follows:
\begin{align}
    \vq = \KRT_{m}[:,{:}3] \mL_{(t,k)}^{{\textrm{3D}}} + \KRT_{m}[:,3] \in \gR^{3\times1}                          , \label{eq:3DP-1}\\
    {\hat{\mL}}_{(m,t,k)} = \left[ \begin{array}{ccc} \vq[0] ~/~ \vq[2] \\ \vq[1] ~/~ \vq[2] \\ \end{array} \right] \in \gR^{2\times1}    , \label{eq:3DP-L}
\end{align}

\noindent\textbf{Loss calculation}.
The SBT supervision comes from the discrepancy between the predicted coordinates $\mL_{(m,t,k)}$ and the estimated coordinates ${\hat{\mL}}_{(m,t,k)}$ from 3D triangulation and reprojection.
The loss function that calculates this discrepancy can vary for regression and heatmap-based detectors. We will introduce our choices in \Secref{sec:supervisionRT-introduction}.

\subsection{Supervision by Registration and Triangulation}\label{sec:supervisionRT-introduction}

{\NAME} leverages supervision from three sources: manually labeled landmarks, registration, and triangulation.
\Figref{fig:high-level} illustrates the procedure of {\NAME} for human pose estimation.
We now detail the loss function for each source of supervision.

\subsubsection{Supervision for Labeled Landmarks}
Most detectors can be categorized into two different types.
(\RomanNumeralCaps{1}) Regression-based model takes an image $\mI$ as input and directly regresses the coordinates of the facial landmarks $\mL$~\cite{lv2017deep,xiong2013supervised}.
(\RomanNumeralCaps{2}) Heatmap-based model predicts for each landmark a heatmap $\mM$, which encodes the confidence of the landmark being found at each location~\cite{newell2016stacked,wei2016convolutional}.
For the regression-based model, an L1 loss was used to measure the error between the prediction $\mL_{(m,t)}$ and ground truth $\mL_{(m,t)}^{*}$:
\begin{align}\label{eq:det-linear}
    \ell_{\textrm{det-L1}} = \sum\nolimits_{k=1}^{K} | \mL_{(m,t,k)} - \mL_{(m,t,k)}^{*} |_{1}.
\end{align}
\noindent For the heatmap-based model, we use the Hourglass model~\cite{newell2016stacked}, and an L2 loss is applied to the predicted heatmap $\mM$ and the ground truth heatmap $\mM^{*}$ for each stage of the Hourglass model. The loss for a single stage is as follows:
\begin{align}\label{eq:det-heatmap}
    \ell_{\textrm{det-L2}} = \sum\nolimits_{k=1}^{K} || \mM_{(m,t,k)} - \mM_{(m,t,k)}^{*} ||_{F}.
\end{align}

\begin{figure}[t!]
\center
\includegraphics[width=\columnwidth]{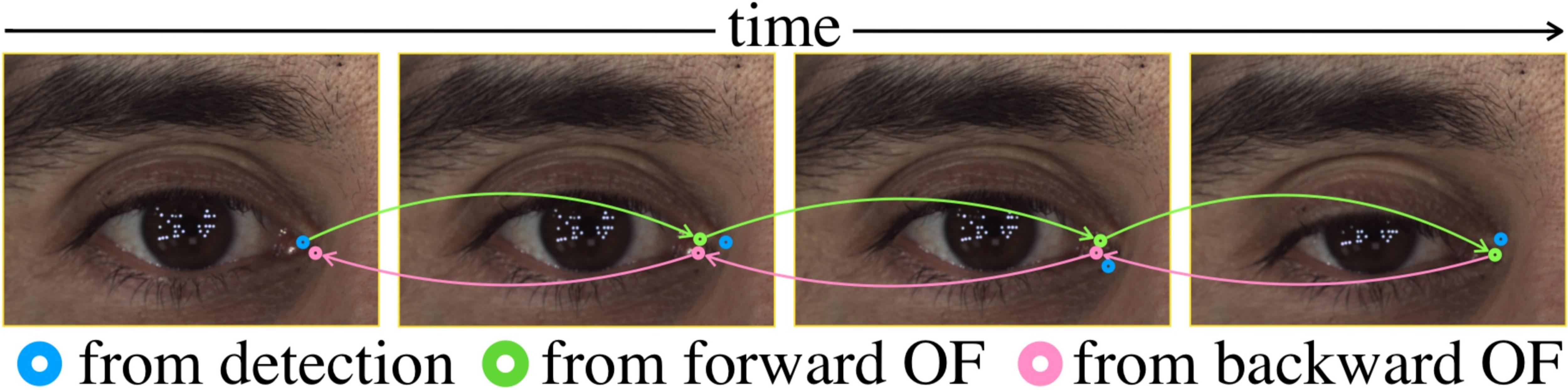}
\caption{
\textbf{Forward-backward communication scheme} between the detector and the OF module during the training procedure.
The green and pink lines indicate the forward and backward OF tracking routes.
The blue/green/pink dots indicate the landmark predictions from the detector/forward-OF/backward-OF.
}
\label{fig:of_check}
\end{figure}

\subsubsection{Supervision-by-Registration}
There are two main points to consider for the loss of SBR: (\RomanNumeralCaps{1}) a suitable function to compare the discrepancy between two detections, and (\RomanNumeralCaps{2}) removing incorrect supervision due to failures in optical flow or landmark detections.
We propose a forward-backward communication scheme between the detection output and the OF module to tackle these two points.
The forward communication computes the discrepancy while the backward communication evaluates the reliability of the OF module.

\noindent
\textbf{Discrepancy comparison.}
The registration loss directly computes the distance between the OF module's predictions ($\widetilde{{\mL}}_{(m,t,k)}$, green dots in~\Figref{fig:of_check}) and the detector's predictions (${\mL}_{(m,t,k)}$, blue dots in~\Figref{fig:of_check}).
The SBR loss for the regression-based model is as follows:
{
\begin{align}\label{eq:sbr-linear}
\ell_{\textrm{sbr-L1}} = \sum\nolimits_{k=1}^{K} \widetilde{\beta}_{(m,t,k)} {||{\mL}_{(m,t,k)}-\widetilde{{\mL}}_{(m,t,k)}||}_{1}.
\end{align}
}\noindent
{$\widetilde{\beta}_{(m,t,k)} \in \{0, 1\}$ indicates the reliability of the $k$-th tracked landmark at time$_t$ in view$_{m}$, which is computed by the backward communication scheme detailed in the next section.}

However, for heatmap-based detectors, we found that soft-$\arg\max$ \cite{dong2018sbr} to extract $(x, y)$ location on heatmaps combined with $\ell_{\textrm{sbr-L1}}$ does not perform well in our experiments (see \Tabref{table:Regression-Mugsy-Objective}).
Therefore, we propose a heatmap-based loss for the heatmap-based detector as follows:
{
\begin{align}\label{eq:sbr-heat}
\ell_{\textrm{sbr-L2}} = \sum\nolimits_{k=1}^{K} \widetilde{\beta}_{(m,t,k)} {||{\mM}_{(m,t,k)}-\widetilde{{\mM}}_{(m,t,k)}||}_{F} ,
\end{align}
}\noindent where ${\mM}_{(m,t,k)}$ indicates the heatmap of the $k$-th landmark at frame$_{(m,t)}$.
$\widetilde{{\mM}}_{(m,t,k)}$ is computed by warping the heatmap from the previous frame (${\mM}_{(m,t-1,k)}$) according to dense optical flow.
$\ell_{\textrm{sbr-L2}}$ can effectively improve the accuracy of heatmap-based detectors.

\noindent
\textbf{Removing incorrect supervision.}
Inaccurate optical flow or landmark detections will lead to inaccurate supervision.
The backward communication scheme, inspired by~\cite{kalal2010forward}, filters out unreliable OF-generated landmarks from the SBR objective.
It consists of three parts:
(\RomanNumeralCaps{1}) a landmark tracked from frame$_{(m,t-1)}$ to frame$_{(m,t)}$ and then from frame$_{(m,t)}$ to frame$_{(m,t-1)}$ should be consistent with its original coordinate. Otherwise, it means the OF module failed for this landmark. More specifically, {$\widetilde{\beta}_{(m,t,k)} = 0$ if $ || \mL_{(m,t-1,k)} - G_{OF}(\mF_{(m,t)},\mF_{(m,t-1)}, \widetilde{{\mL}}_{(m,t,k)}) || > T_{\textrm{FB}} $, where $T_{\textrm{FB}}$ is the threshold for the forward-backward check.}
(\RomanNumeralCaps{2}) the OF-generated results should not be far away from the detection results, i.e., $\widetilde{\beta}_{(m,t,k)} = 0$ if $ || \widetilde{{\mL}}_{(m,t,k)} - {\mL}_{(m,t,k)} || > T_{\textrm{D}} $, where $T_{\textrm{D}}$ is the threshold.
(\RomanNumeralCaps{3}) $\widetilde{\beta}_{(m,t,k)} = 0$ if the coordinates of detection or tracking results is outside the bounding box or image boundary.
If the above three conditions do not hold, we deem the supervision to be reliable and set $\widetilde{\beta}_{(m,t,k)} = 1$.

\subsubsection{Supervision-by-Triangulation}

The SBT loss is calculated as the distance between the detected landmark: ${\mL}_{(m,t,k)}$, and the reprojected landmark: ${\hat{\mL}}_{(m,t,k)}$, as follows:
{
\begin{align}\label{eq:sbt-linear}
\ell_{\textrm{sbt-L1}} = \sum\nolimits_{k=1}^{K} \hat{\beta}_{(m,t,k)} {||{\mL}_{(m,t,k)}-\hat{{\mL}}_{(m,t,k)}||}_{1} .
\end{align}
}\noindent For the heatmap-based model, we can use the soft-$\arg\max$ function to obtain $\mL$ from $\mM$ as in~\cite{dong2018sbr} and then apply \Eqref{eq:sbt-linear}.
However, in experiments, this strategy often leads to unstable optimization (see \Tabref{table:Regression-Mugsy-Objective}). We thus consider the following objective for the heatmap-based model:
{
\begin{align}\label{eq:sbt-heat}
\ell_{\textrm{sbt-L2}} = \sum\nolimits_{k=1}^{K} \hat{\beta}_{(m,t,k)} {||{\mM}_{(m,t,k)}-\hat{{\mM}}_{(m,t,k)}||}_{F} ,
\end{align}
}\noindent where $\hat{{\mM}}_{(m,t,k)}$ is translated from $\mM_{(m,t,k)}$ by the displacement $\hat{{\mL}}_{(m,t,k)} - {\mL}_{(m,t,k)}$.

Similar to SBR, it is also crucial that outliers are removed before applying SBT.
Some projected landmarks result in a very large SBT loss, which suggests that these projected landmarks are likely to be in a wrong location. Using these landmarks in SBT could lead to incorrect supervision.
To remove such outliers, we set $\hat{\beta}_{(m,t,k)} = 0$ if $ || \hat{{\mL}}_{(m,t,k)} - {\mL}_{(m,t,k)} || > T_{\textrm{TRI}} $, where $T_{\textrm{TRI}}$ is the threshold.

\subsubsection{Final Loss Function}
The final loss is a weighted sum of all three kinds of supervisions as follows.
\begin{align}\label{eq:final-loss}
    \ell = \ell_{\textrm{det}} + \omega_{\textrm{sbr}} \ell_{\textrm{sbr}} + \omega_{\textrm{sbt}} \ell_{\textrm{sbt}},
\end{align}
\noindent where $\ell_{\textrm{det}}$, $\ell_{\textrm{sbr}}$, and $\ell_{\textrm{sbt}}$ represent the detection, SBR, and SBT losses.
$\omega_{\textrm{sbr}}$ and $\omega_{\textrm{sbt}}$ are weights of the corresponding losses.

\section{Experimental Analysis}\label{sec:experiments}

\subsection{Datasets}\label{sec:exp-dataset}

To demonstrate the effectiveness of SRT, we performed experiments with 11 datasets summarized in \Tabref{table:dataset}.

\begin{table}[t!]
\setlength{\tabcolsep}{2pt}
\footnotesize
\centering
\caption{
The description of 11 datasets used in our experiments.
``HP'' indicates human pose.
``PF'' and ``PP'' indicate Panoptic-Face and Panoptic-Pose.
}
\begin{tabular}{|c|c|c|c|c|c|c|} \hline\hline
    & modality     &  \makecell{data\\type}   & \#videos  &  \#views  &  \makecell{\#images\\/frames}  & \makecell{\#landm\\-arks}  \\ \hline
\makecell{300-W\\\cite{sagonas2016300,sagonas2013300,sagonas2013semi}} &
      image        & face &     N/A   &    1      &    3837           & 68           \\ \hline
{AFLW}~\cite{koestinger2011annotated} &
      image        & face &     N/A   &    1      &    24386          & 19           \\ \hline
{WFLW}\cite{wu2018look} &
      image        & face &      N/A   &    1     &    10000          & 98           \\ \hline
\makecell{300-VW\\\cite{chrysos2015offline,shen2015first,tzimiropoulos2015project}} &
      video        & face &     114   &    1      &    218597         & 68           \\ \hline
PF~\cite{joo2019panoptic,joo2015panoptic} &
    video          & face &      81   &    7      &    172415         & 70           \\ \hline
VoxCeleb2~\cite{chung2018voxceleb2} &
      video        & face &    29970  &    1      &    2997000        & N/A       \\ \hline
MPII~\cite{andriluka20142d} &
      image        & HP   &     N/A   &    1      &    40522          & 16      \\ \hline
PP~\cite{joo2019panoptic,joo2015panoptic} &
      video        & HP   &    275    &     9     &    825000         & 19      \\ \hline
Human-3.6M~\cite{ionescu2014human3}  &
      video        & HP   &    836    &     4     &    2103096        & 32 \\  \hline
Mugsy-V1  & 
      video        & face &    441    &     6     &    196222         & 18      \\  \hline
Synthetic-Face     &
      image        & face &    N/A    &     1     &   3543            & 18 \\
\hline\hline
\end{tabular}
\label{table:dataset}
\end{table}

\subsubsection{Single-View Image Landmark Detection Datasets}

\textbf{300-W}~\cite{sagonas2016300,sagonas2013300,sagonas2013semi} is a facial landmark detection dataset, which contains more than 3000 facial images with 68 landmarks. We use the common setting on 300-W~\cite{lv2017deep,dong2018sbr,dong2018san}, in which the dataset is split into the training set, the common test set, the challenge test set, and the full test set.

\noindent\textbf{AFLW}~\cite{koestinger2011annotated} is another facial landmark detection dataset, which contains 24386 facial images with 19 landmarks. We follow~\cite{lv2017deep,dong2018sbr} and split the whole dataset into the training set, the frontal test set, and the full test set. We also create two different test sets: one only containing images with yaw degree lower than 30$^{o}$ and another only containing images with yaw degree larger than 60$^{o}$.

\noindent{\textbf{WFLW}~\cite{wu2018look} contains 7500 training images and 2500 test images with 98 facial landmarks.
}

\noindent\textbf{MPII}~\cite{andriluka20142d} is a human pose estimation dataset, containing 28821 training images and 11701 test images. This dataset has annotations for 16 body joints. We use the official training, validation, and test splits.

\noindent{\textbf{Synthetic-Face}~\cite{dong2018sbr} is an internal synthetic face dataset. It contains 1770 training images and 1773 test images from the same person.} This dataset is synthetically generated from a 3D model of a person's face. As a result, we can guarantee the annotations of this dataset have zero error, thus by perturbing these accurate annotations, we can measure the effect of inaccuracies in annotations. Sample images are shown in the supplementary material.

\subsubsection{Single-View Video Landmark Detection Datasets}

\noindent\textbf{300-VW}~\cite{chrysos2015offline,shen2015first,tzimiropoulos2015project} contains 50 training videos with 95192 frames. It provides three test sets: A has 31 videos with 62135 frames, B has 19 videos with 32805 frames, C has 14 videos with 26338 frames.

\noindent\textbf{VoxCeleb2}~\cite{chung2018voxceleb2} contains thousands of videos of people talking. It does not have landmark annotation, and we thus use it as an unlabeled video dataset for enhancing detectors with SBR.
Since it is very large, we only sample 5 videos for each identity and extract the first 100 frames of each video.

\begin{table*}[t!]
\centering
\footnotesize
\setlength{\tabcolsep}{2pt}
\caption{
We show the effect of utilizing various unlabeled data sets to enhance the regression-based detector.
We report NME and {\PERR} on three test subsets of 300-W.
We also report NME, AUC@0.08, and {\PERR} on 300-VW A, B, and C.
$\WS{sbr}$ and $\WS{sbt}$ indicate the loss weights of SBR and SBT supervision, respectively.
$\times$ denotes the corresponding supervision is not used.
For all experiments in this table, we use the 49 landmarks, excluding landmarks on the facial boundary.
The SBR/SBT/SRT numbers are averaged over 3 runs.
}
\begin{tabular}{| c | c | c | c | c | c | c | c | c | c | c | c | c | c | c | c |} \hline\hline
   \multirow{2}{*}{\makecell{Labeled\\Data}} & \multirow{2}{*}{\makecell{Unlabeled\\Data}}  &  \multirow{2}{*}{$\WS{sbr}$} & \multirow{2}{*}{$\WS{sbt}$} & \multicolumn{3}{c|}{300-W Test Set} & \multicolumn{3}{c|}{300-VW A} & \multicolumn{3}{c|}{300-VW B} & \multicolumn{3}{c|}{300-VW C} \\\cline{5-16}
  & & & & Common  &  Challenge  &  Full  &  NME   &  AUC & {\PERR} & NME   &  AUC & {\PERR} & NME   &  AUC & {\PERR} \\\hline
 \multirow{22}{*}{\makecell{300-W\\Training\\Set}} &
      N/A              &    $\times$     &      $\times$    & 2.99 / 12.08\XTH & 5.83 / 21.44\XTH & 3.55 / 13.97\XTH & 4.58 & 53.76 & 15.27\XTH & 3.98 & 52.81 & 15.46\XTH & 9.31 & 45.90 & 24.03\XTH \\
\cline{2-16}
   &  300-VW           &        0.1      &      $\times$    & 2.86 / 10.73\XTH & 5.56 / 19.82\XTH & 3.39 / 12.39\XTH & 4.39 & 55.00 & 13.76\XTH & 3.76 & 54.48 & 13.84\XTH & 9.19 & 48.63 & 21.90\XTH \\
   &  300-VW           &        0.5      &      $\times$    & 2.86 / 9.99\XTH  & 5.55 / 19.88\XTH & 3.39 / 11.47\XTH & 4.12 & 57.43 & 12.59\XTH & 3.61 & 56.00 & 12.51\XTH & 8.47 & 52.57 & 19.97\XTH \\
   &  300-VW           &        1.0      &      $\times$    & 2.87 / 9.74\XTH  & 5.50 / 19.79\XTH & 3.39 / 11.15\XTH & 4.10 & 57.94 & 12.36\XTH & 3.62 & 56.22 & 12.12\XTH & 8.53 & 53.12 & 19.51\XTH \\
   &  300-VW           &        2.0      &      $\times$    & 2.93 / 9.67\XTH  & 5.59 / 18.54\XTH & 3.45 / 11.06\XTH & 4.16 & 57.45 & 12.31\XTH & 3.67 & 55.41 & 11.75\XTH & 8.39 & 53.45 & 19.10\XTH \\
\cline{2-16}
   &  VoxCeleb2        &        0.1      &      $\times$    & 2.87 / 10.49\XTH & 5.46 / 18.25\XTH & 3.37 / 11.78\XTH & 4.39 & 53.97 & 13.29\XTH & 3.88 & 52.80 & 13.56\XTH & 8.64 & 48.55 & 20.66\XTH \\
   &  VoxCeleb2        &        0.5      &      $\times$    & 2.85 / 10.04\XTH & 5.43 / 16.92\XTH & 3.36 / 11.16\XTH & 4.32 & 54.80 & 12.94\XTH & 3.85 & 53.07 & 12.90\XTH & 8.61 & 47.15 & 18.77\XTH \\
   &  VoxCeleb2        &        1.0      &      $\times$    & 2.86 / 9.91\XTH  & 5.58 / 17.29\XTH & 3.39 / 11.05\XTH & 4.34 & 53.86 & 13.33\XTH & 3.87 & 52.79 & 13.16\XTH & 8.61 & 43.17 & 18.88\XTH \\
   &  VoxCeleb2        &        2.0      &      $\times$    & 2.90 / 10.43\XTH & 5.69 / 18.13\XTH & 3.44 / 11.65\XTH & 4.59 & 51.80 & 15.26\XTH & 3.94 & 51.83 & 14.38\XTH & 9.28 & 35.81 & 20.88\XTH \\
\cline{2-16}
   &   PF             &        0.1      &      $\times$    & 2.87 / 10.73\XTH  & 5.49 / 20.24\XTH & 3.38 / 12.23\XTH & 4.25 & 55.22 & 13.27\XTH & 3.76 & 53.96 & 13.54\XTH & 8.70 & 50.09 & 21.12\XTH \\
   &   PF             &        0.5      &      $\times$    & 2.87 / 10.58\XTH  & 5.59 / 19.76\XTH & 3.40 / 12.17\XTH & 4.27 & 55.33 & 13.78\XTH & 3.83 & 53.46 & 14.32\XTH & 8.50 & 50.18 & 22.76\XTH \\
   &   PF             &        1.0      &      $\times$    & 2.88 / 10.43\XTH  & 5.58 / 21.06\XTH & 3.41 / 12.05\XTH & 4.35 & 54.71 & 14.47\XTH & 3.92 & 52.33 & 15.28\XTH & 8.90 & 45.75 & 24.88\XTH \\
   &   PF             &        2.0      &      $\times$    & 2.95 / 10.81\XTH  & 5.89 / 25.90\XTH & 3.53 / 12.90\XTH & 5.60 & 50.76 & 18.20\XTH & 4.48 & 49.19 & 19.35\XTH & 15.82 & 24.54 & 37.69\XTH \\
\cline{2-16}
   &   PF             &     $\times$    &         0.1      & 2.86 / 10.84\XTH  & 5.39 / 19.31\XTH & 3.35 / 12.12\XTH & 4.19 & 54.60 & 13.24\XTH & 3.79 & 53.86 & 13.58\XTH & 8.59 & 48.02 & 21.23\XTH \\
   &   PF             &     $\times$    &         0.5      & 2.86 / 10.63\XTH  & 5.39 / 18.23\XTH & 3.35 / 11.93\XTH & 4.08 & 55.72 & 12.09\XTH & 3.76 & 53.85 & 12.79\XTH & 8.85 & 48.56 & 19.10\XTH \\
   &   PF             &     $\times$    &         1.0      & 2.88 / 10.31\XTH  & 5.47 / 17.98\XTH & 3.38 / 11.68\XTH & 4.11 & 55.21 & 11.60\XTH & 3.87 & 52.56 & 12.51\XTH & 8.77 & 46.35 & 18.60\XTH \\
   &   PF             &     $\times$    &         2.0      & 2.94 / 10.70\XTH  & 5.53 / 18.17\XTH & 3.45 / 12.11\XTH & 4.25 & 54.01 & 11.65\XTH & 3.99 & 51.19 & 12.78\XTH & 9.22 & 40.27 & 18.49\XTH \\
\cline{2-16}
   &   PF             &         0.1     &         0.1      & 2.85 / 10.56\XTH  & 5.42 / 19.59\XTH & 3.36 / 11.93\XTH & 4.15 & 55.41 & 12.85\XTH & 3.77 & 54.20 & 13.42\XTH & 8.21 & 49.91 & 20.18\XTH \\
   &   PF             &         0.5     &         0.5      & 2.87 / 10.13\XTH  & 5.49 / 18.03\XTH & 3.39 / 11.72\XTH & 4.09 & 55.90 & 11.88\XTH & 3.83 & 53.04 & 12.83\XTH & 8.44 & 46.60 & 19.12\XTH \\
   &   PF             &         0.5     &         1.0      & 2.90 / 10.45\XTH  & 5.52 / 19.12\XTH & 3.41 / 11.75\XTH & 4.17 & 54.70 & 11.79\XTH & 3.95 & 51.72 & 13.06\XTH & 8.78 & 42.61 & 18.71\XTH \\
   &   PF             &         1.0     &         1.0      & 2.92 / 10.36\XTH  & 5.72 / 19.15\XTH & 3.47 / 11.99\XTH & 4.36 & 52.80 & 12.00\XTH & 4.09 & 50.31 & 13.45\XTH & 9.31 & 34.13 & 19.78\XTH \\
   &   PF             &         2.0     &         2.0      & 3.01 / 11.20\XTH  & 6.56 / 23.45\XTH & 3.71 / 13.16\XTH & 6.68 & 45.89 & 18.48\XTH & 5.49 & 46.55 & 17.54\XTH & 18.55 & 18.75 & 30.24\XTH \\
\hline\hline
\end{tabular}
\label{table:Regression-300W-ALL}
\end{table*}

\subsubsection{Multi-view Landmark Detection Datasets}

\noindent\textbf{CMU-Panoptic}~\cite{joo2019panoptic} is collected from a massively synchronized and geometrically calibrated multi-view system, which contains more than 30 high-definition views (1920$\times$1080 resolution).
In this dataset, they provide 3D facial landmark labels for some videos.
By only keeping frames where the frontal face is visible and choosing the first 2000 frames of each video, we construct a new dataset named Panoptic-Face (PF).
Similar to Panoptic-Face, we use the multi-view videos with labeled human pose information as a new dataset for pose estimation, named Panoptic-Pose.

\noindent\textbf{Human-3.6M}~\cite{ionescu2014human3} is collected in a space with the size of 3 meters $\times$ 4 meters. 4 views from 4 different synchronized cameras are used.
We use the subjects 1, 5, 6, 7, 8, 9, 11 as the unlabeled multi-view videos to enhance the detector.

\noindent\textbf{Mugsy-V1} is an internal multi-view facial video dataset.
It contains 6 views and 441 videos in total.
There are 196222 frames in total, and 12543 frames are annotated with 18 facial landmarks.
We use 6394 annotated frames as the test set, and the remaining 6149 annotated frames together with 183679 unlabeled frames as the training set.
Note that (\RomanNumeralCaps{1}) all annotated frames are in the same view, and (\RomanNumeralCaps{2}) the training frames are not from the same video as any test frames.
When we validate the hyper-parameters, we randomly select 50\% training samples for the validation set.

\subsection{Experiment Settings}\label{sec:exp-setting}

\noindent\textbf{Data Augmentation.}
We use the same data augmentation for most experiments unless otherwise specified.
Given a bounding box, we sequentially apply six steps.
(\RomanNumeralCaps{1}) Expand this box by 20\% height and width.
(\RomanNumeralCaps{2}) Randomly resize this box in the range of [90\%, 110\%] with probability of 50\%.
(\RomanNumeralCaps{3}) Randomly translate this box, in which the maximum displacement of the box center is 10\% height or width.
(\RomanNumeralCaps{4}) Randomly rotate this box by the maximum 40 degrees.
(\RomanNumeralCaps{5}) Crop a face/body image by this box and convert this image into gray-scale.
(\RomanNumeralCaps{6}) Apply intensity augmentation on this cropped image by multiplying one random scalar in [0.6, 1.4] to each channel.

\begin{table}[t!]
\centering
\setlength{\tabcolsep}{4pt}
\caption{
Results of SBR and SBT to enhance the regression-based detector on Mugsy-V1.
$\times$ denotes the corresponding supervision is not used. Note that since Mugsy-V1 includes an unlabeled portion in the training set, no other unlabeled datasets are used to train the detectors.
}
\begin{tabular}{| c | c | c | c | c | c | c | c |} \hline\hline
\multirow{2}{*}{$\WS{sbr}$} & \multirow{2}{*}{$\WS{sbt}$}  & \multicolumn{3}{c|}{Validation} & \multicolumn{3}{c|}{Test} \\\cline{3-8}
  &  &    NME     &    AUC   & $P\textrm{-}error$  &    NME   &   AUC    &  $P\textrm{-}error$ \\\hline
     $\times$   &    $\times$   &    4.18    &   48.79  &  4.96\XTH   &   4.63   &   43.47  &  5.14\XTH  \\
\hline
     0.1        &    $\times$   &    4.11    &   49.62  &  5.07\XTH   &   4.53   &   44.45  &  5.06\XTH  \\
     0.5        &    $\times$   &    4.09    &   50.01  &  4.91\XTH   &   4.48   &   45.10  &  4.86\XTH  \\
     1.0        &    $\times$   &    4.05    &   50.37  &  4.77\XTH   &   4.44   &   45.48  &  4.71\XTH  \\
     2.0        &    $\times$   &    4.07    &   50.43  &  4.78\XTH   &   4.47   &   45.17  &  4.82\XTH  \\
\hline
   $\times$     &       0.1     &    4.06    &   49.67  &  4.51\XTH   &   4.48   &   44.94  &  4.60\XTH  \\
   $\times$     &       0.5     &    4.02    &   50.21  &  3.82\XTH   &   4.39   &   45.92  &  3.80\XTH  \\
   $\times$     &       1.0     &    4.01    &   50.25  &  3.42\XTH   &   4.42   &   45.48  &  3.38\XTH  \\
   $\times$     &       2.0     &    4.14    &   48.71  &  3.09\XTH   &   4.60   &   43.39  &  3.05\XTH  \\
\hline
     0.5        &       0.5     &   3.85     &   52.02  &  3.45\XTH   &   4.36   &   46.30  &  3.68\XTH  \\
     1.0        &       0.5     &   3.85     &   52.00  &  3.47\XTH   &   4.38   &   46.14  &  3.70\XTH  \\
     2.0        &       0.5     &   3.87     &   51.87  &  3.40\XTH   &   4.38   &   46.14  &  3.69\XTH  \\
     0.5        &       1.0     &   3.90     &   51.51  &  3.17\XTH   &   4.43   &   45.49  &  3.39\XTH  \\
     1.0        &       1.0     &   3.92     &   51.21  &  3.25\XTH   &   4.42   &   45.61  &  3.33\XTH  \\
     2.0        &       1.0     &   3.91     &   51.38  &  3.09\XTH   &   4.44   &   45.39  &  3.33\XTH  \\
\hline\hline
\end{tabular}
\label{table:Regression-Mugsy}
\end{table}

\begin{table*}[t!]
\centering
\footnotesize
\setlength{\tabcolsep}{4pt}
\caption{
We utilize various unlabeled data sets to enhance the regression-based detector on different AFLW test sets.
yaw$\leq$30$^{o}$: all test faces with the yaw degree lower than 30 degrees.
yaw$\geq$60$^{o}$: all test faces with the yaw degree higher than 60 degrees.
Our approach significantly improves the precision (reducing {\PERR}) compared to the baseline models.
The SBR/SBT/SRT numbers are averaged over 3 runs.
}
\begin{tabular}{| c | c | c | c | c | c | c | c | c | c | c | c | c | c | c | c |} \hline\hline
  \multirow{2}{*}{\makecell{Labeled\\Data}} &  \multirow{2}{*}{\makecell{Unlabeled\\Data}} & \multirow{2}{*}{$\WS{sbr}$} & \multirow{2}{*}{$\WS{sbt}$} & \multicolumn{3}{c|}{AFLW-Front}  & \multicolumn{3}{c|}{AFLW-Full}   & \multicolumn{3}{c|}{yaw$\leq$30$^{o}$}   & \multicolumn{3}{c|}{yaw$\geq$60$^{o}$} \\\cline{5-16}
  & & & & NME &  AUC  & {\PERR} & NME &  AUC  & {\PERR} & NME &  AUC  & {\PERR} & NME &  AUC  & {\PERR} \\\hline
  \multirow{22}{*}{\makecell{AFLW\\Training\\Set}}         &   N/A     & $\times$ & $\times$ & 1.89 & 77.25 & 19.49\XTH & 2.65 & 68.07 & 12.98\XTH & 2.12 & 74.07 & 14.82\XTH & 3.94 & 53.94 & 31.30\XTH \\\cline{2-16}
      &  300-VW   & 0.1      & $\times$ & 1.80 & 77.78 & 18.10\XTH & 2.58 & 68.69 & 11.93\XTH & 2.07 & 74.58 & 13.53\XTH & 3.83 & 54.64 & 30.05\XTH \\
      &  300-VW   & 0.5      & $\times$ & 1.81 & 77.64 & 17.65\XTH & 2.58 & 68.60 & 11.35\XTH & 2.08 & 74.48 & 13.10\XTH & 3.86 & 54.49 & 29.93\XTH \\
      &  300-VW   & 1.0      & $\times$ & 1.82 & 77.54 & 17.52\XTH & 2.60 & 68.49 & 11.21\XTH & 2.08 & 74.39 & 12.90\XTH & 3.88 & 54.31 & 29.65\XTH \\
      &  300-VW   & 2.0      & $\times$ & 1.85 & 77.21 & 18.06\XTH & 2.62 & 68.18 & 12.14\XTH & 2.11 & 74.06 & 13.59\XTH & 3.89 & 53.95 & 29.18\XTH \\
\cline{2-16}
      & VoxCeleb2 & 0.1      & $\times$ & 1.81 & 77.74 & 17.85\XTH & 2.58 & 68.64 & 11.78\XTH & 2.07 & 74.53 & 13.45\XTH & 3.82 & 54.58 & 29.42\XTH \\
      & VoxCeleb2 & 0.5      & $\times$ & 1.81 & 77.65 & 17.88\XTH & 2.58 & 68.59 & 12.02\XTH & 2.07 & 74.49 & 13.60\XTH & 3.81 & 54.44 & 29.14\XTH \\
      & VoxCeleb2 & 1.0      & $\times$ & 1.83 & 77.49 & 18.87\XTH & 2.60 & 68.42 & 14.63\XTH & 2.10 & 74.35 & 15.23\XTH & 3.84 & 54.14 & 28.65\XTH \\
      & VoxCeleb2 & 2.0      & $\times$ & 1.86 & 77.09 & 19.63\XTH & 2.64 & 67.96 & 15.60\XTH & 2.14 & 73.91 & 15.87\XTH & 3.88 & 53.60 & 28.97\XTH \\
\cline{2-16}
      &   PF     & 0.1      & $\times$ & 1.81 & 77.70 & 18.13\XTH & 2.57 & 68.62 & 11.97\XTH & 2.07 & 74.49 & 13.67\XTH & 3.81 & 54.49 & 29.74\XTH \\
      &   PF     & 0.5      & $\times$ & 1.82 & 77.59 & 17.79\XTH & 2.59 & 68.48 & 11.53\XTH & 2.09 & 74.36 & 13.32\XTH & 3.85 & 54.39 & 29.35\XTH \\
      &   PF     & 1.0      & $\times$ & 1.83 & 77.49 & 17.81\XTH & 2.61 & 68.26 & 11.71\XTH & 2.10 & 74.24 & 13.45\XTH & 3.85 & 54.12 & 29.40\XTH \\
      &   PF     & 2.0      & $\times$ & 1.88 & 76.97 & 18.65\XTH & 2.67 & 67.63 & 12.32\XTH & 2.14 & 73.72 & 14.16\XTH & 3.96 & 53.21 & 30.89\XTH \\
\cline{2-16}
      &   PF     & $\times$ &   0.1    & 1.82 & 77.69 & 18.16\XTH & 2.58 & 68.67 & 11.97\XTH & 2.08 & 74.55 & 13.76\XTH & 3.84 & 54.59 & 29.90\XTH \\
      &   PF     & $\times$ &   0.5    & 1.83 & 77.54 & 17.88\XTH & 2.59 & 68.51 & 11.50\XTH & 2.09 & 74.36 & 13.31\XTH & 3.83 & 54.48 & 29.19\XTH \\
      &   PF     & $\times$ &   1.0    & 1.84 & 77.39 & 17.90\XTH & 2.62 & 68.21 & 11.54\XTH & 2.11 & 74.11 & 13.45\XTH & 3.90 & 54.13 & 29.38\XTH \\
      &   PF     & $\times$ &   2.0    & 1.87 & 77.05 & 18.76\XTH & 2.67 & 67.68 & 12.28\XTH & 2.15 & 73.74 & 14.14\XTH & 3.97 & 53.17 & 30.71\XTH \\
\cline{2-16}
      &   PF     & 0.1      &   0.1    & 1.81 & 77.68 & 17.92\XTH & 2.58 & 68.66 & 11.70\XTH & 2.07 & 74.48 & 13.50\XTH & 3.83 & 54.65 & 29.72\XTH \\
      &   PF     & 0.5      &   0.5    & 1.84 & 77.37 & 17.78\XTH & 2.61 & 68.29 & 11.59\XTH & 2.11 & 74.17 & 13.51\XTH & 3.85 & 54.27 & 28.95\XTH \\
      &   PF     & 0.5      &   1.0    & 1.84 & 77.31 & 18.17\XTH & 2.62 & 68.10 & 11.82\XTH & 2.11 & 74.02 & 13.78\XTH & 3.90 & 53.86 & 29.93\XTH \\
      &   PF     & 1.0      &   1.0    & 1.85 & 77.19 & 18.34\XTH & 2.65 & 67.91 & 12.15\XTH & 2.12 & 73.95 & 13.88\XTH & 3.95 & 53.62 & 30.06\XTH \\
      &   PF     & 2.0      &   2.0    & 1.90 & 76.64 & 19.21\XTH & 2.71 & 67.18 & 12.73\XTH & 2.17 & 73.33 & 14.69\XTH & 4.03 & 52.62 & 30.85\XTH \\\hline
 \hline
\end{tabular}
\label{table:Regression-AFLW}
\end{table*}

\noindent\textbf{The Setup of Regression-based Model.}
Due to the high efficiency of MobileNet-V2~\cite{sandler2018mobilenetv2}, we choose it as the backbone of our regression-based detector.
We made some modification based on MobileNet-V2 and its final layer is an fully connected layer, which outputs a vector with the dimension of $2\times{K}$.
In all experiments, we train this regression-based detector via Adam~\cite{kingma2015adam}.
We train the detector for 200 epochs with the initial learning rate as 0.001.
We reduce the learning rate by a factor of 10 at 100-th and 150 epochs.
For facial datasets, we resize the input face into 96$\times$96 by default, but use 240$\times$320 for Mugsy-V1.
We found that using the restart technique~\cite{loshchilov2017sgdr} can significantly improve the performance of the regression-based model.
Specifically, we train the model for 200 epochs, and then we increase the learning rate to 0.1 and retrain the model for another 200 epochs.
We repeat this procedure twice on 300-W and AFLW, and repeat it ten times on Mugsy-V1.

\noindent\textbf{The Setup of Heatmap-based Model.}
We use the stacked hourglass model~\cite{newell2016stacked} as our heatmap-based detector. We use four hourglass stages and set the recursive step as 3. The number of channels of the intermediate features is 256. We follow the official training strategy introduced in ~\cite{newell2016stacked}.
We use the RMSProp optimizer and an initial learning rate of 0.001.
For face datasets, we train the model for 160 epochs and decay the learning rate by a factor of 2 at the 60-th, 90-th, 110-th, and 130-th epoch.
We resize the input face into 256$\times$256 for the heatmap-based detector.

\noindent
\textbf{Training Procedure.}
We assume the bounding box for face or human is already obtained.
During training, the input is always a cropped face for facial landmark detection or human body for pose estimation.
We utilized a stage wise training strategy with two steps:
(1) optimizing the detector with detection loss only;
and (2) enhancing the detector with detection, SBR, and SRT losses jointly.

\noindent\textbf{Training with SBR.}
We perform the OF tracking over three consecutive frames by default.
For differential LK OF, the $\Omega$ in \Eqref{eq:optical-flow} is a $13\times13$ patch centered at the landmark.
The maximum iterations of OF is 20 and the convergence threshold for $\Delta{\bf p}=10^{-6}$.
For the input feature of the OF module, we use the gray-scale image.
In the forward-backward communication scheme, $T_{\textrm{FB}}$ and $T_{\textrm{D}}$ are both 1\% of the square root of the area of the face/body bounding box.
When using the bilinear interpolation OF, we use the Farneback algorithm~\cite{farneback2003two} to compute dense optical flow.

\noindent\textbf{Training with SBT.}
We perform 3D triangulation on four views, i.e., $M=4$.
Specifically, given an image in one view, we first randomly sample three different views from the current view. We use the corresponding landmark predictions in these four views to obtain the 3D coordinates, and then project 3D coordinates to 2D coordinates in these four views.
The threshold $T_{\textrm{TRI}}$ is 1\% of the square root of the area of the face/body bounding box.

\noindent\textbf{Balanced sampling of labeled and unlabeled data per batch.}
The amount of unlabeled data is oftentimes much larger than the amount of labeled data.
Therefore, we explicitly ensure that the network is seeing a balanced amount of labeled and unlabeled samples, such that the network does not ``forget'' human supervision.
Specifically, in one batch, we have 32 labeled images, 32 video triplets (each contains three consecutive frames), and 16 multi-view quadruplets (each contains images from four different views).

\noindent\textbf{Evaluating accuracy}.
We measured the Normalized Mean Error (NME) and the Area Under the Curve (AUC)@0.08 metric.
We use the inter-ocular distance on 300-W and the face size on AFLW for normalization.
When reporting NME and AUC, we omit the \% notation.
For pose estimation, we use the standard Percentage of Correct Keypoints by a fraction of the head size (PCKh) for evaluation.

\noindent\textbf{Evaluating precision}.
We propose to utilize the Equivariant Landmark Transformation (ELT)~\cite{honari2018improving}, which was originally used as a loss function, as a proxy to measure the precision of the detector.
ELT applies a known affine transformation to an image, and we would expect the landmark detections on both the original image and the transformed image to just differ by the affine transformation. 
If the detections do not completely follow the affine transformation, this is a sufficient condition for the detector not being very precise,
i.e., the detections are not consistent across images.
Therefore, we use this as a proxy to measure the precision of the detector.
More specifically, given an image $\mI$, we apply two separate random data augmentation transformations\footnote{We apply three steps: (\RomanNumeralCaps{1}) randomly scale the bounding box in a range of [0.8, 1.2], (\RomanNumeralCaps{2}) randomly translate the box with the maximum displacement being 10\% height horizontally and 10\% width vertically, (\RomanNumeralCaps{3}) randomly rotate the box with the maximum degree of 30.} and obtain $\mI^{a}$ and $\mI^{b}$.
This procedure can be formulated as $\mI^{a}=\textrm{Affine}_{\Theta^{a}}(\mI)$ and $\mI^{b}=\textrm{Affine}_{\Theta^{b}}(\mI)$, where $\textrm{Affine}_{\Theta^{a}}$ denotes the affine transformation parametrized by $\Theta^{a}$. 
By giving $\mI^{a}$ and $\mI^{b}$ into the detector, we obtain landmark predictions $\mL^{a}_p$ and $\mL^{b}_p$ respectively for landmark $p$. We then compute the following metric to evaluate precision:
\begin{align}
    P\textrm{-}error = \frac{1}{\eta} \sum\nolimits_{k} ||\textrm{Affine}^{-1}_{\Theta^{a}}(\mL^{a}_{k}) - \textrm{Affine}^{-1}_{\Theta^{b}}(\mL^{b}_{k})|| , \label{eq:p-error}
\end{align}
\noindent where {\PERR} (Precision-error) is calculated as the mean discrepancy of each point pair $\mL^{a}_{p}$ and $\mL^{b}_{p}$ once the inverse affine transform has been applied to the points. $\eta$ is a normalization constant, which is the square root of the area of the bounding box.
The advantage of ELT is that no additional labeled data is necessary, thus {\PERR} is not affected by the inconsistencies in human annotations.

\begin{table}[t!]
\centering
\setlength{\tabcolsep}{3pt}
\caption{
The effect of different loss functions for detectors on Mugsy-V1.
}
\begin{tabular}{| c | c | c | c | c | c | c |} \hline\hline
\multirow{2}{*}{Losses} &  \multicolumn{3}{c|}{Validation} & \multicolumn{3}{c|}{Test} \\\cline{2-7}
                            &  NME   &  AUC          & {\PERR}  & NME   &  AUC          & {\PERR}\\\hline\hline
    \multicolumn{7}{|c|}{the regression-based detector}     \\\hline
     $-$                 & 4.18   &  48.79           & 4.96\XTH & 4.63   &  43.47   & 5.14\XTH \\\hline
L1 on $\mL$ for SBR      & 4.05   &  50.37           & 4.77\XTH & 4.44   &  45.48   & 4.71\XTH \\
L2 on $\mL$ for SBR      & 4.13   &  49.40           & 5.02\XTH & 4.53   &  44.65   & 5.14\XTH \\\hline
L1 on $\mL$ for SBT      & 4.01   &  50.25           & 3.42\XTH & 4.42   &  45.48   & 3.38\XTH \\
L2 on $\mL$ for SBT      & 4.08   &  49.58           & 4.75\XTH & 4.49   &  44.90   & 4.70\XTH \\\hline
L1 on $\mL$ for SRT      & 3.85   &  52.02           & 3.45\XTH & 4.36   &  46.30   & 3.68\XTH \\
\hline\hline
    \multicolumn{7}{|c|}{the heatmap-based detector}         \\\hline
    $-$                  & 3.78   &  52.78         & 2.65\XTH &  4.04  &  50.05           &  2.60\XTH \\\hline
L1 on $\mL$ for SBR      & 6.37   &  21.83         & 3.29\XTH &  7.67  &  12.12           &  7.12\XTH \\
L2 on $\mL$ for SBR      & 3.95   &  51.79         & 2.93\XTH &  3.97  &  50.88           &  2.33\XTH \\
L2 on $\mM$ for SBR      & 3.77   &  52.85         & 2.35\XTH &  3.96  &  51.02           &  2.31\XTH \\\hline
L1 on $\mL$ for SBT      & 5.36   &  33.60         & 0.79\XTH &  5.54  &  31.77           &  0.73\XTH \\
L2 on $\mL$ for SBT      & 3.78   &  52.85         & 2.29\XTH &  3.98  &  50.87           &  2.20\XTH \\
L2 on $\mM$ for SBT      & 3.77   &  52.98         & 2.24\XTH &  3.80  &  50.88           &  2.20\XTH \\\hline
L2 on $\mM$ for SRT      & 3.76   &  52.98         & 2.10\XTH &  3.94  &  51.20           &  2.08\XTH \\
\hline\hline
\end{tabular}
\label{table:Regression-Mugsy-Objective}
\end{table}

\subsection{Ablation Studies}\label{sec:exp-ablation}

We did a series of experiments to study the effect of (\RomanNumeralCaps{1}) different kinds of optical flow algorithms for approximation.
(\RomanNumeralCaps{2}) different weights for $\WS{sbr}$ and $\WS{sbt}$;
(\RomanNumeralCaps{3}) different kinds of supervision;
(\RomanNumeralCaps{4}) different kinds of data source;
(\RomanNumeralCaps{5}) different kinds of loss;
(\RomanNumeralCaps{6}) annotation noise;
and (\RomanNumeralCaps{7}) effectiveness of removing incorrect supervision.
Since the regression-based detector is more efficient and simple than the heatmap-based detector, we use the regression-based detector for most ablation studies.

\noindent\textbf{The effect of different approximation methods for OF.}
As discussed in \Secref{sec:method-approx-of}, we could approximate OF at sub-pixel locations through bilinear interpolation on a pre-computed flow field.
We empirically compare the difference between the actual OF results and the interpolation results.
On the first 200 frames of all 14 videos in the 300-VW C test set, we use the ground truth coordinate of each landmark from the previous frame as initial location, and compute the tracked results at the current frame.
There is only 0.02 average pixel discrepancy between the interpolated results and the actual OF results.
This shows that the OF computed from bilinear interpolation is very accurate.
We further evaluated seven different OF methods (please see supplementary material for details).
Different OF methods yield similar performance, but the differentiable Lucas Kanade OF~\cite{dong2018sbr} is significantly slower than utilizing bilinear approximation (e.g., more than 41 times slower than Farneback~\cite{farneback2003two}). Since the differentiable Lucas Kanade OF takes too much time, we leverage the interpolation strategy for our remaining experiments.
We choose Farneback~\cite{farneback2003two} as our default OF algorithm because (1) it is one of the most popular methods, (2) it achieves similar performance than others, (3) its speed is acceptable.

\noindent\textbf{The effect of different loss weights: $\WS{sbr}$, $\WS{sbt}$.}
We tested different loss weights on Mugsy-V1 in \Tabref{table:Regression-Mugsy}.
Among $\{$0.1, 0.5, 1.0, 2.0$\}$ for $\WS{sbr}$,  $\WS{sbr}$ of 1.0 gives the highest AUC and lowest NME on the validation set. Different $\WS{sbr}$ gives similar {\PERR}.
Among $\{$0.1, 0.5, 1.0, 2.0$\}$ for $\WS{sbt}$, a higher $\WS{sbt}$ will result in a lower {\PERR}, which means the model is precise, i.e., more robust to spatial transformation.
However, the training AUC drops when $\WS{sbt}$ becomes too large, i.e., the detectors becomes more precise but also more inaccurate. 
This could be due to the triangulation constraint forcing the detector to predict a landmark at a consistent-in-3D but wrong location.
Therefore, in order to balance both NME, AUC, and {\PERR}, we use $\WS{sbr}=1.0$ for SBR, and  $\WS{sbt}=1.0$ for SBT. For SRT (SBR+SBT), we use $\WS{sbr}=0.5$, $\WS{sbt}=0.5$.

\begin{table}[t!]
\setlength{\tabcolsep}{2.1pt}
\centering
\caption{
The comparison of the normalized mean error (NME) on AFLW.
``HB'' indicates the heatmap-based detector.
}
\begin{tabular}{|c|c c c|c|c|} \hline\hline
Methods   & CCL~\cite{zhu2016unconstrained} & SAN~\cite{dong2018san} & Wing~\cite{feng2018wing} & HB & HB + SRT  \\ \hline
AFLW-Full & 2.72                            & 1.91          &  1.47 & 2.31 & 2.26 \\ \hline
AFLW-Front& 2.17                            & 1.85          & -     & 1.71 & 1.64 \\\hline\hline
\end{tabular}
\label{table:AFLW-SOTA}
\end{table}

\begin{table}[t!]
\setlength{\tabcolsep}{2pt}
\centering
\caption{
PCKh with a threshold of 0.5 evaluation metric on the MPII Human Pose validation set.
}
\begin{tabular}{| l | c | c | c | c |} \hline\hline
      Methods                             &  SRT            & Params (M) & PCKh  & {\PERR}    \\\hline
CU-Net~\cite{tang2018cu}                  &  $-$            & 24.18      & 71.51 & 119.21\XTH \\
CPM~\cite{wei2016convolutional}           &  $-$            &  25.40     & 72.12 & 92.27\XTH  \\
MSPN~\cite{li2019rethinking}              &  $-$            & 26.06      & 76.30 & 89.39\XTH  \\\hline
heatmap-based~\cite{newell2016stacked}   &  $-$            & 25.26      & \textbf{80.50} & 84.42\XTH  \\
heatmap-based                            &  Panoptic-Pose  & 25.26      & 80.03 & 83.17\XTH  \\
heatmap-based                            &  Human-3.6M     & 25.26      & 80.15 & \textbf{83.05\XTH}  \\
\hline\hline
\end{tabular}
\label{table:MPII-SOTA}
\end{table}

\noindent\textbf{Comparing SBR and SBT.}
\Tabref{table:Regression-300W-ALL} shows SBR and SBT results on 300-W.
Several conclusions can be made:
(\RomanNumeralCaps{1}) temporal information from 300-VW, VoxCeleb2, and Panoptic-Face can improve the detector's accuracy.
(\RomanNumeralCaps{2}) multi-view information from Panoptic-Face can improve the detector's accuracy.
(\RomanNumeralCaps{3}) the mutual benefit from both temporal and multi-view cues can further enhance the detection performance.
(\RomanNumeralCaps{4}) using the multi-view cue of Panoptic-Face obtains lower NME on the image test set than using the temporal cue.
\Tabref{table:Regression-AFLW} and \Tabref{table:Regression-Mugsy} shows the results on AFLW and Mugsy-V1 respectively.
We can observe similar phenomena as \Tabref{table:Regression-300W-ALL}.
\Tabref{table:Regression-Mugsy} also shows that multi-view information is more beneficial to precision improvement than temporal information.

\begin{table}[t!]
\setlength{\tabcolsep}{3pt}
\footnotesize
\centering
\caption{
The comparison of NME w.r.t. the inter-ocular distance on 300-W.
We also compute {\PERR} on the full test set of 300-W.
}
\begin{tabular}{|l|c|c|c|c|} \hline\hline
      Methods                            & Common & Challenging & Full Set   & {\PERR}  \\\hline
      \multicolumn{5}{|c|}{68 landmarks} \\\hline
      Two-Stage \cite{lv2017deep}        & 4.36   & 7.42        & 4.96    &   -     \\
 Pose-Invariant~\cite{jourabloo2017pose} & 5.43   & 9.88        & 6.30    &   -     \\
 PCD-CNN~\cite{kumar2018disentangling}   & 3.67   & 7.62        & 4.44    &   -     \\
      SBR~\cite{dong2018sbr}             & 3.28   & 7.58        & 4.10    &   -     \\
      SAN~\cite{dong2018san}             & 3.34   & 6.60        & 3.98    & 17.41   \\
      LAB~\cite{wu2018look}              & 2.98   & 5.19        & 3.49    &   -     \\
      ODN~\cite{zhu2019robust}           & 3.56   & 6.67  & 4.17 & - \\\hline
 heatmap-based                           & 2.81   & 5.50        & 3.34    & 16.15    \\
 heatmap-based + ELT                     & 2.81   & 5.53        & 3.33    & 16.01    \\
 heatmap-based + SRT                 & 2.80   & 5.61        & 3.39    & 15.23    \\\hline
 \multicolumn{5}{|c|}{49 landmarks (without landmarks on the facial contour)} \\\hline
 SAN~\cite{dong2018san}                  & 2.42   & 5.42        & 3.01    & 13.10 \\\hline
 regression-based                        & 2.99   & 5.83        & 3.55    & 13.97 \\
 regression-based + SRT              & 2.84   & 5.36        & 3.31    & 10.89 \\
 heatmap-based                           & 2.00   & 3.93        & 2.38    & 9.94   \\
 heatmap-based + SRT                 & 1.98   & 3.99        & 2.41    & 8.87   \\
\hline\hline
\end{tabular}
\label{table:300-W-ALL-SOTA}
\end{table}

\noindent\textbf{The effect of different kinds of the unlabeled data source.}
We try different unlabeled datasets in \Tabref{table:Regression-300W-ALL} and \Tabref{table:Regression-AFLW}.
All four different unlabeled datasets can enhance the base detector, while there are some interesting observations.
(\RomanNumeralCaps{1}) Unlabeled 300-VW provides the best NME and AUC results on 300-VW A test category compared to others.
(\RomanNumeralCaps{2}) Even if VoxCeleb2 is 10 times larger than 300-VW, ``300-W + VoxCeleb2'' obtains a worse result than ``300-W + 300-VW''. The faces in VoxCeleb2 are blurred and are different from 300-VW. In contrast, the faces in the 300-VW training set are somewhat similar to that of the 300-VW test set w.r.t. pose, motion, etc. 
(\RomanNumeralCaps{3}) Unlabeled Panoptic-Face does not improves the detector as much as 300-VW or VoxCeleb2. This might be due to the limited diversity (number of identities) of Panoptic-Face.

\noindent\textbf{The effect of different kinds of the losses} is analyzed in \Tabref{table:Regression-Mugsy-Objective}.
We test different kinds loss functions when enhancing a detector with SBR and SBT.
When using the regression-based detector, both ``L1 on $\mL$'' and ``L2 on $\mL$'' (L1, L2 loss on coordinates) can significantly improve the base detector.
L1-based SBR and SBT obtain a better performance compared to L2-based SBR and SBT, thus we selected ``L1 on $\mL$'' for regression-based detector for SBR and SBT.

When using the heatmap-based detector, there are three different loss functions: ``L1 on $\mL$'', ``L2 on $\mL$'', and ``L2 on $\mM$'' (heatmaps).
SBR and SBT with ``L1 on $\mL$'' do not work well\footnote{In \Tabref{table:Regression-Mugsy-Objective}, ``L1 on $\mL$'' for SBT may have very good {\PERR}, but NME and AUC are very poor. This is an example of a detector being precise but inaccurate.}.
``L2 on $\mM$'' gives a slightly lower NME and higher AUC than ``L2 on $\mL$''.
Therefore, we choose the loss ``L2 on $\mM$'' for the heatmap-based detector.

\begin{figure}[t]
\center
\includegraphics[width=\linewidth]{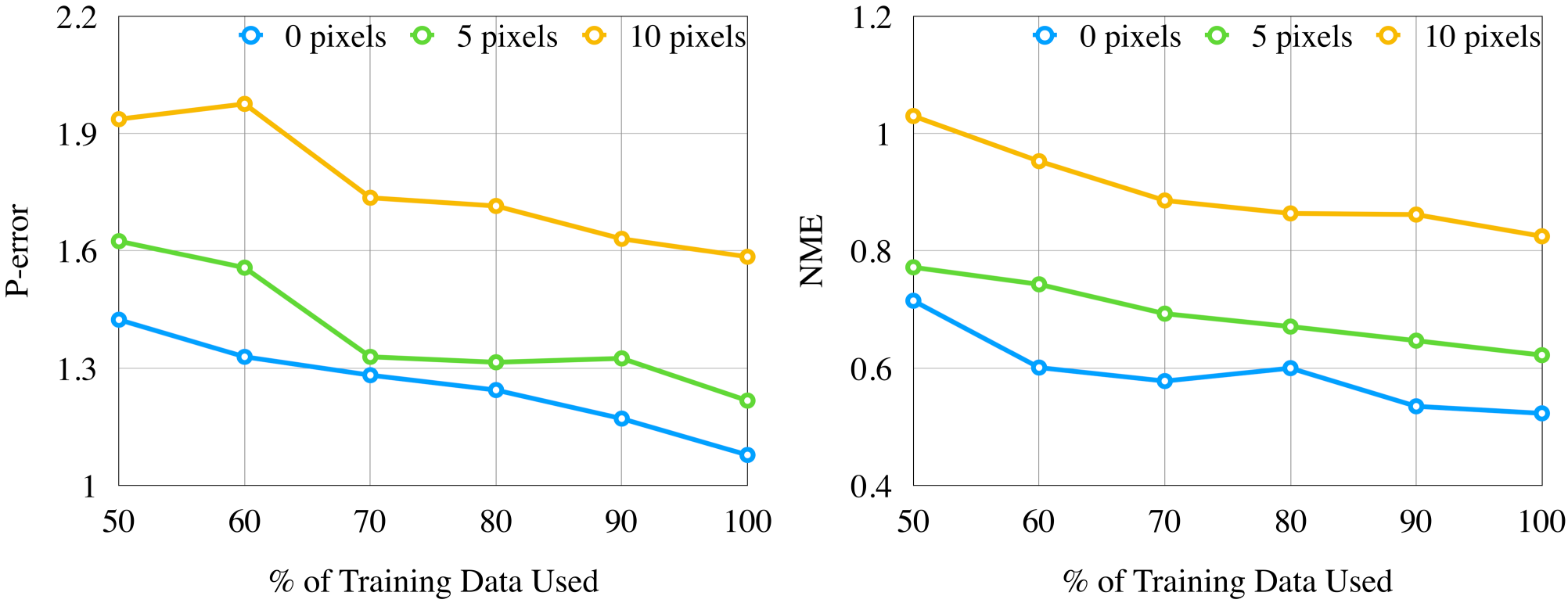}
\caption{
The effect of data size and noise level for the heatmap-based model on the test set of Synthetic-Face.
We randomly add Gaussian noise with std=0,5,10 pixels to the ground truth labels of the training set.
}
\label{fig:Sync-Effect}
\end{figure}

\noindent\textbf{The effect of inaccurate annotations.}
We perturb the training data of Synthetic-Face with various amounts of Gaussian noise, and train the heatmap-based detector 3 times with different random seeds on varying amounts of training data (50\% to 100\%) and noise levels (standard deviation 0, 5, and 10).
Results shown in \Figref{fig:Sync-Effect} suggests the following:
(\RomanNumeralCaps{1}) Training with more data can improve both precision and accuracy.
(\RomanNumeralCaps{2}) Training with more accurate labels can improve both precision and accuracy. For example, having 100\% of the training data with std=10 pixel noise leads to worse performance than having just 50\% of the training data but no error in labels.
(\RomanNumeralCaps{3}) NME and {\PERR} are highly correlated. The advantage of {\PERR} is that it does not require annotations thus will not be affected by inaccurate annotations.
This makes it a good proxy for performance when testing set accuracy has saturated.

\begin{table}[t!]
\setlength{\tabcolsep}{2pt}
\centering
\caption{
Results on three 300-VW test sets. Two different sources of bounding box detections are used as input.
HB + SRT means heatmap-based detector trained with SRT.
Most results come from~\cite{chrysos2018comprehensive,deng2019joint}.
}
\begin{tabular}{|c c | c c | c c | c c |} \hline\hline
\multicolumn{2}{|c|}{Methods} & \multicolumn{2}{c|}{300-VW A} & \multicolumn{2}{c|}{300-VW B} & \multicolumn{2}{c|}{300-VW C} \\\hline
  Detection & Landmark            & AUC  & Failure & AUC  & Failure & AUC  & Failure \\\hline
\multirow{3}{*}{DPM} & CFSS~\cite{zhu2015face}       & 76.6 & 3.74    & 77.0 & 1.32   & 72.4 & 5.23   \\
                     & ERT~\cite{kazemi2014one}      & 77.7 & 3.44    & 77.2 & 1.51   & 72.1 & 6.08   \\
                     & HB + SRT                      & 78.1 & 2.32    & 77.8 & 1.31   & 74.0 & 4.98\\\hline 
\multirow{3}{*}{MTCNN} & CFSS~\cite{zhu2015face}     & 73.4 & 8.51    & 72.5 & 8.52   & 72.6 & 5.69    \\
                       & STCSR~\cite{yang2015facial} & 79.1 & 2.40    & 78.8 & 0.32   & 71.0 & 4.46 \\
                                             & HB + SRT  & 74.3 & 7.37 & 73.7 & 7.16 & 73.2 & 4.91 \\\hline
\end{tabular}
\label{table:X-300-VW-SOTA}
\end{table}

\noindent\textbf{Effectiveness of removing incorrect supervision.}
Here, we quantitatively show how much supervision are regarded as incorrect and removed.
The percentage of $\widetilde{\beta}_{(m,t,k)}$ and $\hat{\beta}_{(m,t,k)}$ that are set to 0, i.e. supervision removed, varies based on the datasets. We show an example of SRT enhancing the regression-based detector on AFLW. When using 300-VW with SBR, the zero percentage of $\widetilde{\beta}_{(m,t,k)}$ is about 3.16\%. When using Panoptic-Face with SBR, the zero percentage of $\widetilde{\beta}_{(m,t,k)}$ is about 8.95\%.
When using Panoptic-Face with SBT only, the zero percentage is about 32.91\%.
When using Panoptic-Face with SRT, zero percentage of $\widetilde{\beta}_{(m,t,k)}$ and $\hat{\beta}_{(m,t,k)}$ are about 9.15\% and 51.91\%, respectively.

However, the removal of incorrect supervision is not always perfect.
One common failure case is false positives in the forward-backward consistency check.
For a landmark, if the detector makes the same mistake on two consecutive frames, and if the optical flow between the two detections are also consistent, then this will pass the check and be incorrectly used as supervision in SRT.
To quantitatively measure the frequency of these failure cases, we run SRT with our 300-W base detector on 300-VW and leverage the available per-frame labels of facial-landmarks to measure the failure rate. 
Landmarks that pass the forward-backward consistency check but have more than 0.05 normalized distance w.r.t. the ground truth labels are considered as false positives.
Results show that 0.46\% landmarks suffer from false positives in the consistency check, i.e., our method can filter out more than 99\% of the incorrect supervision.

\begin{table}[t!]
\setlength{\tabcolsep}{1.3pt}
\centering
\caption{
Comparison results on WFLW~\cite{wu2018look}. ``PF'' indicates Panoptic-Face.
}
\begin{tabular}{|c| c c | c c c c |} \hline\hline
\multicolumn{3}{|c|}{Methods}  & NME  & AUC@0.1 & Failure@0.1 & {{\PERR}} \\\hline
\multicolumn{3}{|c|}{LAB~\cite{wu2018look}} & 5.27 & 53.23   & 7.56        & 17.21  \\ \hline
base model          & SBR   & SBT & NME  & AUC@0.1 & Failure@0.1 & {{\PERR}} \\\hline
regression-based    & $-$   & $-$ & 6.72 & 41.88   & 14.16       & 22.16     \\
regression-based    & 300-VW & $-$ & 6.55 & 43.00   & 13.76       & 13.97     \\
regression-based    & PF    & $-$ & 6.64 & 41.94   & 14.24       & 20.35     \\
regression-based    & PF    & PF  & 6.54 & 42.61   & 13.48       & 20.27     \\\hline
heatmap-based       & $-$   & $-$ & 5.33 & 53.04   & 8.44        & 18.05     \\
heatmap-based       & 300-VW & $-$ & 5.13 & 54.59   & 7.28        & 16.36     \\
heatmap-based       & PF    & $-$ & 5.13 & 54.65   & 7.08        & 15.49     \\
heatmap-based       & PF    &  PF & 5.13 & 54.64   & 7.07        & 15.01     \\
\hline
\end{tabular}
\label{table:SOTA-WFLW}
\end{table}

\begin{figure*}[t]
\center
\includegraphics[width=\textwidth]{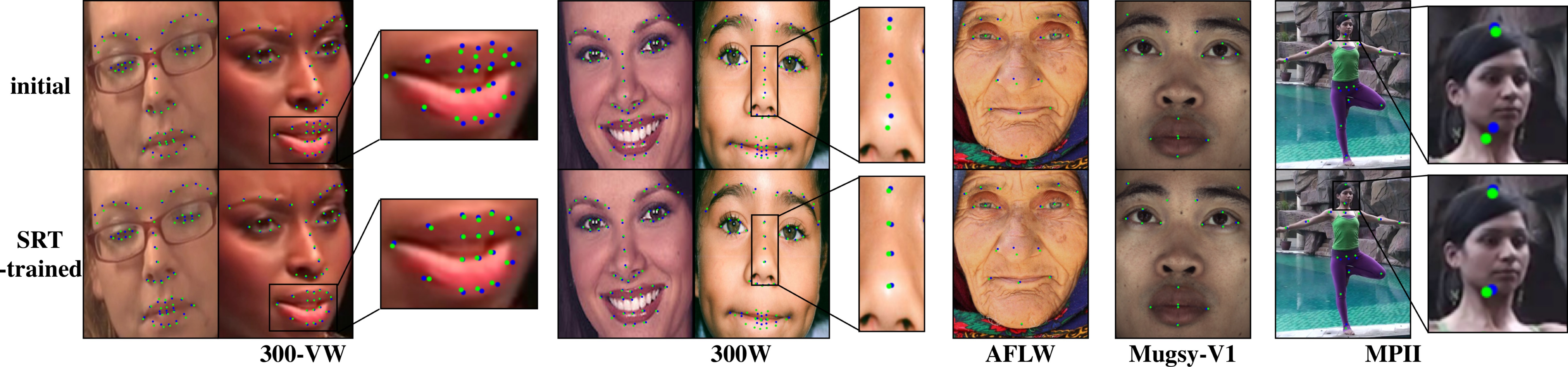}
\caption{
Visualization results of the regression-based models on 300-W, 300-VW, AFLW, and Mugsy-V1, and the heatmap-based models on MPII.
Green and blue points are prediction results and ground truth labels, respectively.
}
\label{fig:results-final-SRT}
\end{figure*}

\subsection{Comparison with the State-of-the-art}\label{sec:exp-sota}

We compare our algorithm with other state-of-the-art algorithms on both facial and body landmark detection.
For facial landmark detection, the unlabeled data used for SRT are from the 300-VW training set, VoxCeleb2 and Panoptic-Face.
For body landmark detection, the unlabeled data used for SRT are from Panoptic-Pose and Human-3.6M.

\noindent\textbf{Results on 300-W} are shown in \Tabref{table:300-W-ALL-SOTA}.
We ran SRT on 68 landmarks of 300-W, which harms the heatmap detector. This is because some landmarks, such as facial contour, are ambiguous across different views.
By filtering out such outlier, we observe a significant improvement based on the regression-based detector.
There is still no clear improvement when SRT is applied to the heatmap-based detector.

\noindent\textbf{Results on 300-VW}.
We compare with some baseline methods from~\cite{chrysos2018comprehensive} in \Tabref{table:X-300-VW-SOTA}.
For each face bounding box detection algorithm used, our method 
is competitive with state-of-the-art methods.

\noindent\textbf{Results on AFLW} are shown in \Tabref{table:AFLW-SOTA}.
SRT improves the heatmap-based detector by about 4\% w.r.t. NME for AFLW-Front.

\noindent\textbf{Results on WFLW}.
We show that SRT can significantly improve the precision of both regression-based and heatmap-based detectors in \Tabref{table:SOTA-WFLW}.
They also slightly reduce the NME and increase the AUC.
With the assistance of SRT, we achieve competitive results compared to LAB~\cite{wu2018look}.

\noindent\textbf{Results on MPII}.
We show results of our SRT on the MPII pose estimation dataset in \Tabref{table:MPII-SOTA}, which reports PCKh@0.5 of the single crop evaluation setting on the validation set.
For a fair comparison, we tune some structure hyper-parameters of different models to make the number of parameters similar. Unfortunately, the performance drops compared to the base detector.
There could be two reasons: (\RomanNumeralCaps{1}) hyper-parameters should be carefully tuned on different datasets, such as SBR/SBT weights and forward-backward communication thresholds;
and (\RomanNumeralCaps{2}) the diversity of unlabeled multi-view videos is not enough to obtain a good performance, since there are only nine unlabeled multi-view videos in Panoptic-Pose.

\noindent\textbf{Discussion}.
For heatmap-based detectors, SRT improved performance on Mugsy-V1 (\Tabref{table:Regression-Mugsy-Objective}), WFLW (\Tabref{table:SOTA-WFLW}), and AFLW (\Tabref{table:AFLW-SOTA}), while performance did not improve on MPII (\Tabref{table:MPII-SOTA}) and 300-W (\Tabref{table:300-W-ALL-SOTA}).
We hypothesize that the incorrect supervision that was not removed by our checks (see \Secref{sec:exp-ablation}) could have led to the drop in performance.
The heatmap-based detector has powerful representation ability, thus it could remember both (\RomanNumeralCaps{1}) the good supervision from the labeled data and (\RomanNumeralCaps{2}) the bad supervision from the failure cases of registration and triangulation which our filtering method could not filter out. 
Thus the generalization ability is reduced by those failure cases leading to degraded performance on the test set.
On the other hand, SRT still improves the heatmap detectors performance on AFLW, WFLW, and Mugsy-V1. One reason could be because the number of training images in AFLW and WFLW is much larger than that of 300-W, thus leading to a better initial landmark location prediction as input to SRT.
In sum, SRT may not always lead to accuracy improvements, but we still see consistent improvement in precision.

\section{Conclusion and Future Work}\label{sec:conclusion}

In this manuscript, we propose SRT, an unsupervised method to improve a regression or heatmap-based landmark detection model by leveraging registration and triangulation supervision, which does not require any additional manual annotations.
Our conclusions are as follows.
(\RomanNumeralCaps{1}): both registration and triangulation can improve accuracy and precision, and they complement each other.
(\RomanNumeralCaps{2}): there is no clear winner to which supervision, registration or triangulation, is more effective.
(\RomanNumeralCaps{3}): bilinear optical flow interpolation for registration works as well as differentiable Lucas Kanade optical flow~\cite{dong2018sbr} while providing much faster training speed.
(\RomanNumeralCaps{4}): the unlabeled data used should have a similar distribution as the labeled training and testing data, and large amounts of out-of-domain data is not as effective as smaller amounts of in-domain data.
(\RomanNumeralCaps{5}): false positives of the forward-backward check is a common way that SRT might degrade the detector's performance. This is due to the SRT losses striving for consistency, i.e. precision, which might not necessarily coincide with the correctness of the detection, i.e. accuracy.
(\RomanNumeralCaps{6}): {\PERR} could be a good proxy for performance especially when testing set accuracy has saturated, as {\PERR} is not affected by inconsistencies in annotations. However, a low {\PERR} does not necessarily mean good accuracy, so {\PERR} should not be used in a vacuum.

In order to further push this research forward, one key ingredient missing is a large-scale, high resolution, multi-camera dataset of faces and bodies with high quality annotations for a large number of subjects under different environments. The current multi-view landmark detection datasets have a large number of video frames but a small number of subjects. Such limited subjects are insufficient for both robust training and convincing evaluation. Also, investigating how to achieve high annotation quality is also crucial in both training and evaluation of detectors. In addition to the dataset, we suggest designing new evaluation protocols for landmark detection to test a detector's robustness w.r.t. the quality of images, geometric calibration, and so on. While the accuracy of landmark detection has been boosted again and again, little effort has been devoted to the study of robustness.

\bibliographystyle{IEEEtran}
\bibliography{IEEEabrv,egbib}

\begin{IEEEbiography}[{\includegraphics[width=1in,height=1.25in,clip,keepaspectratio]{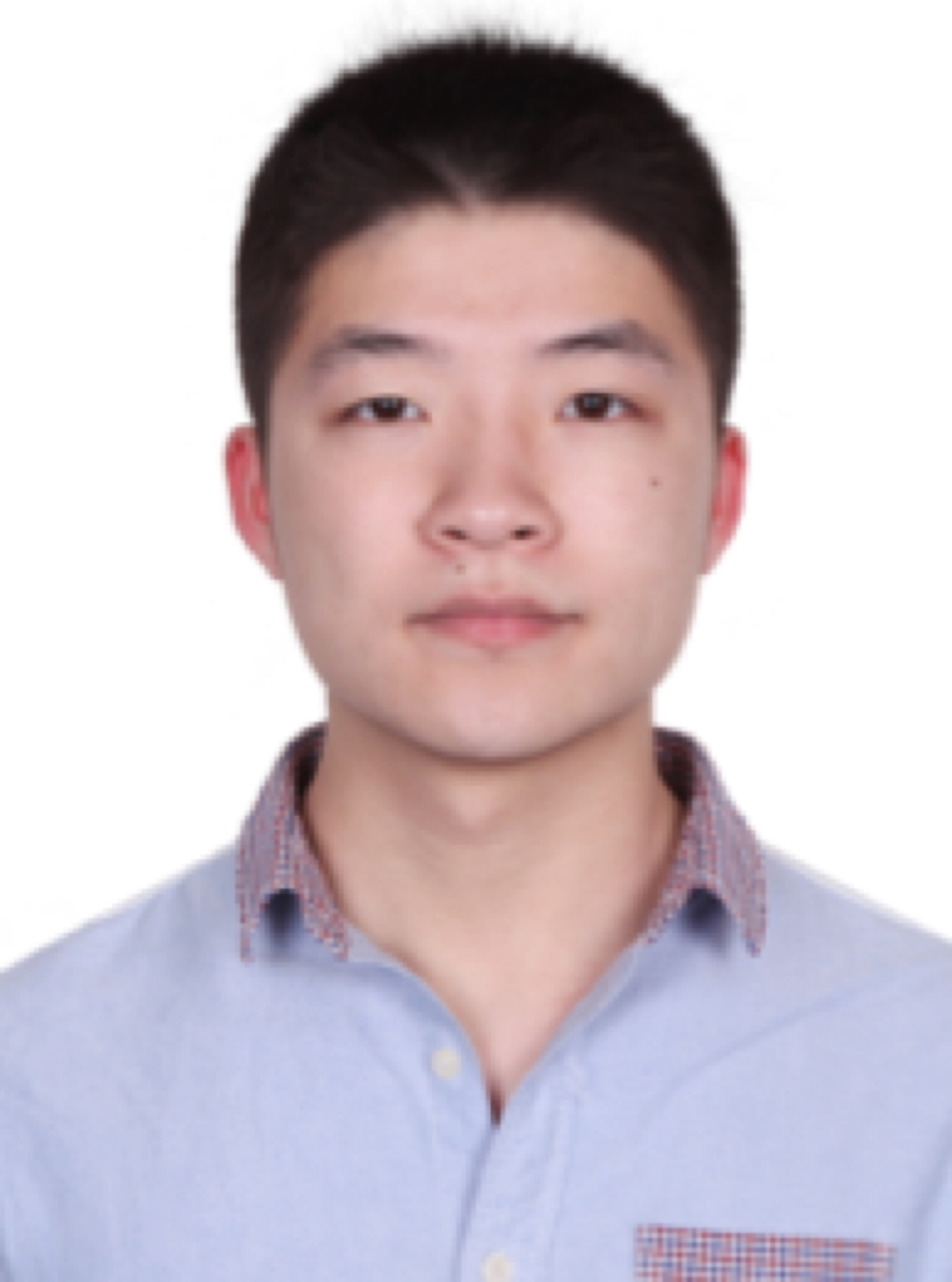}}]{Xuanyi Dong} received the B.E. degree in Computer Science and Technology from Beihang University, Beijing, China, in 2016.
He is currently a Ph.D. student at at School of Computer Science, University of Technology Sydney, Australia.
His research interests include automated deep learning and its application to real world applications.
\end{IEEEbiography}

\begin{IEEEbiography}[{\includegraphics[width=1in,height=1.25in,clip,keepaspectratio]{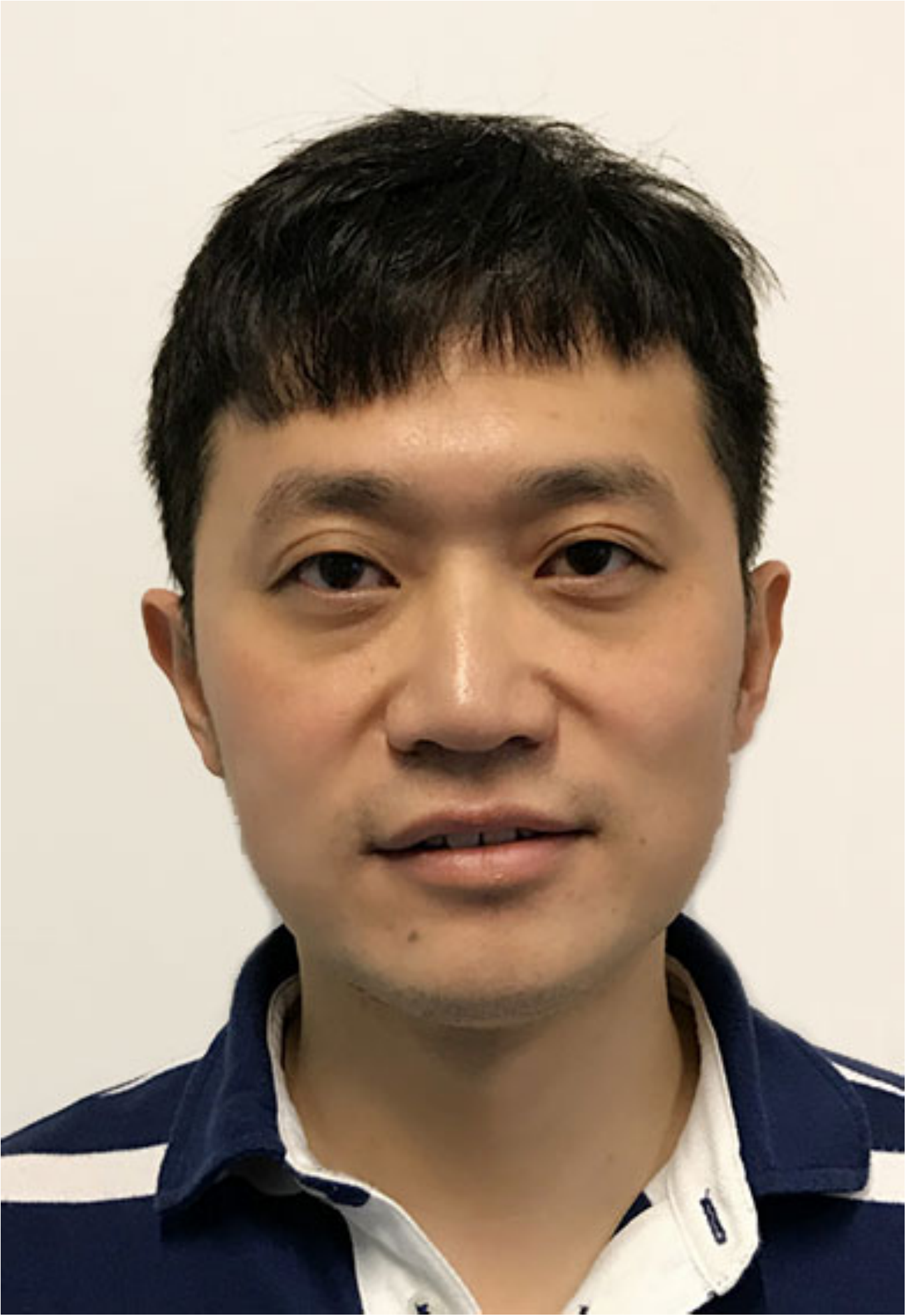}}]{Yi Yang} received the Ph.D. degree in computer science from Zhejiang University, Hangzhou, China, in 2010. He was a post-doctoral researcher with the School of Computer Science, Carnegie Mellon University, Pittsburgh, PA, USA.
He is currently a Professor with University of Technology Sydney, Australia. His current research interest includes machine learning and its applications to multimedia content analysis and computer vision.
\end{IEEEbiography}

\begin{IEEEbiography}[{\includegraphics[width=1in,height=1.25in,clip,keepaspectratio]{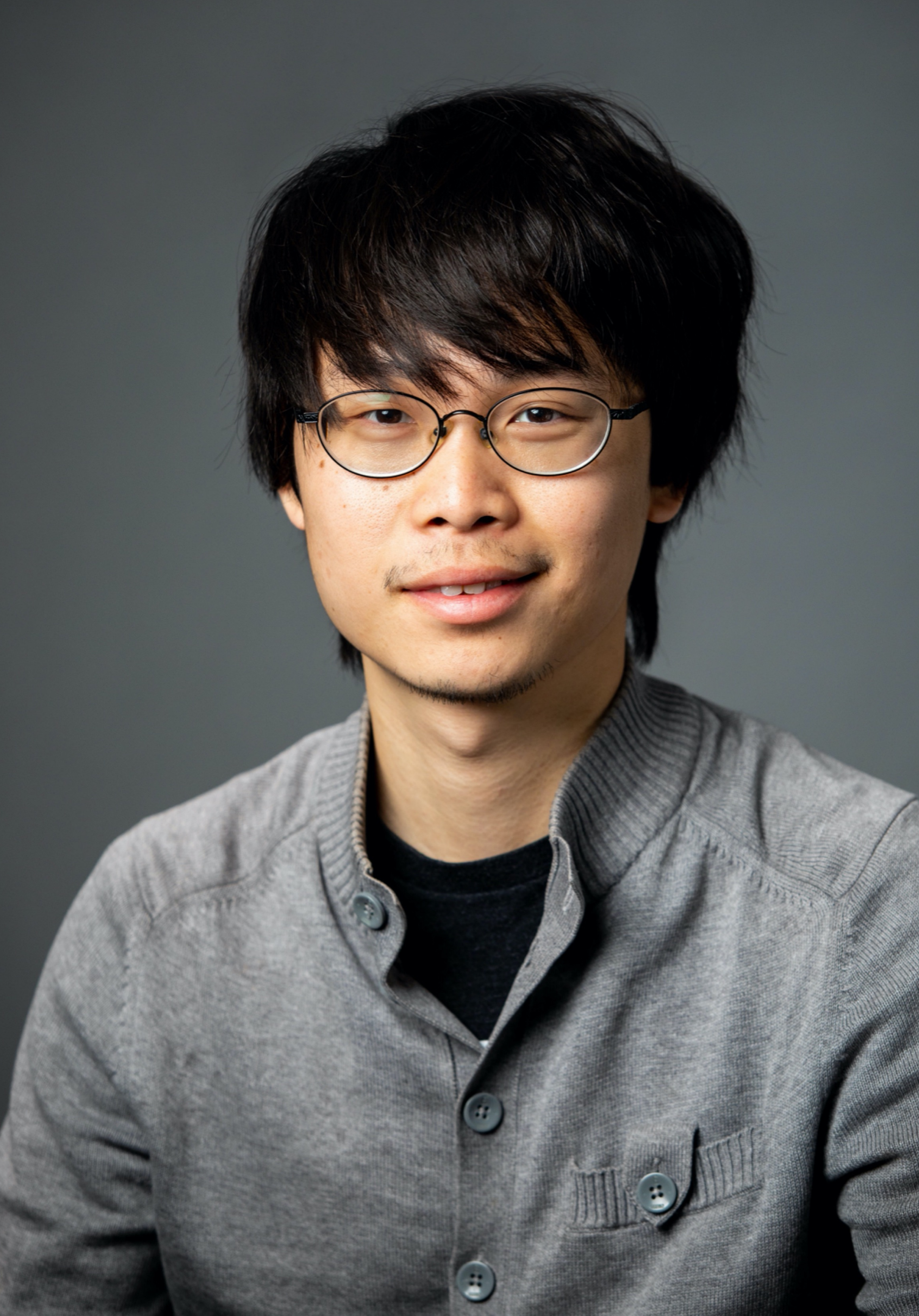}}]{Shih-En Wei} is currently a research scientist focused on machine learning and computer vision for social VR in Facebook Reality Labs, Pittsburgh, Pennsylvania. Prior to that, he received the M.S. in Robotics from the School of Computer Science, Carnegie Mellon University, Pittsburgh. He also received his B.S. in Electrical Engineering and M.S. degree in Communication Engineering from National Taiwan University, Taipei, Taiwan. His research interests mainly lie in tracking human's behavior including eyes, face, hands, and body poses.
\end{IEEEbiography}

\begin{IEEEbiography}[{\includegraphics[width=1in,height=1.25in,clip,keepaspectratio]{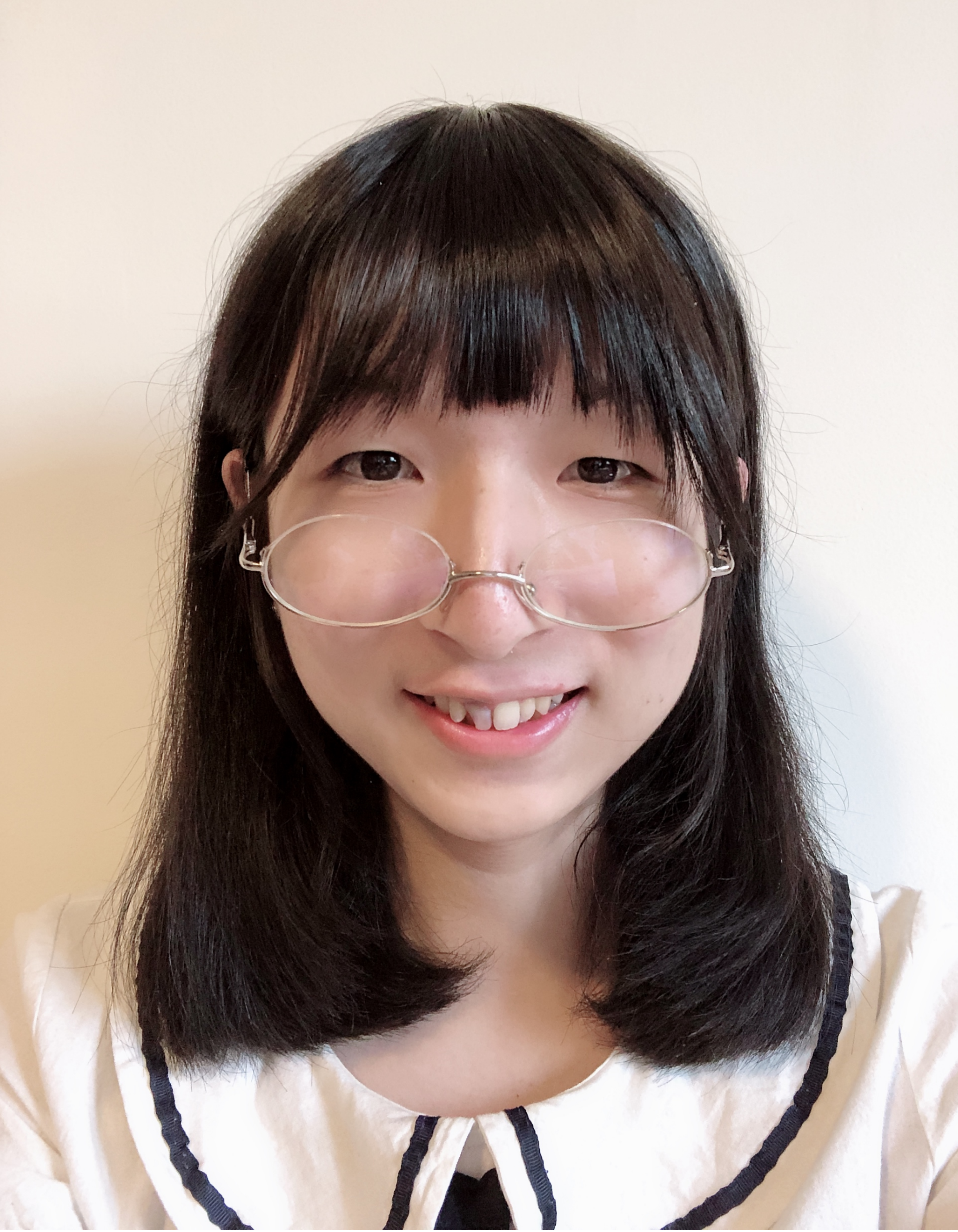}}]{Xinshuo Weng} received the BSc degree from Wuhan University, China, and the MS degree from Carnegie Mellon University. She is currently a Ph.D. student at the Robotics Institute of Carnegie Mellon University. Before starting her Ph.D. program, she was working at Oculus Research Pittsburgh (now Facebook Reality Lab) as a research engineer. Her research interests include computer vision and machine learning, with a special interest in 3D vision and self-supervised learning.
\end{IEEEbiography}

\begin{IEEEbiography}[{\includegraphics[width=1in,height=1.25in,clip,keepaspectratio]{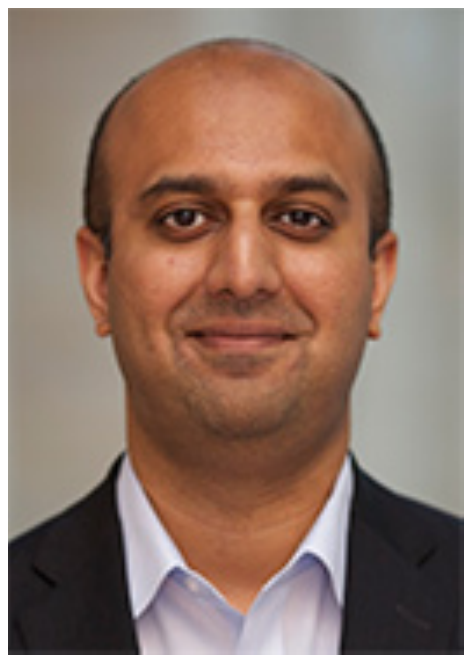}}]{Yaser Sheikh} directs the Facebook Reality Lab in Pittsburgh, which is focused on achieving metric telepresence in AR and VR, and is an adjunct professor at Carnegie Mellon University. His research broadly focuses on machine perception and rendering of social behavior, spanning sub-disciplines in computer vision, computer graphics, and machine learning. With colleagues and students, he has won the Honda Initiation Award (2010), Popular Science’s "Best of What’s New" Award, best student paper award at CVPR (2018), best paper finalist awards at (CVPR 2019), best paper awards at WACV (2012), SAP (2012), SCA (2010), ICCV THEMIS (2009), best demo award at ECCV (2016), and he received the Hillman Fellowship for Excellence in Computer Science Research (2004). Yaser has served as a senior committee member at leading conferences in computer vision, computer graphics, and robotics including SIGGRAPH (2013, 2014), CVPR (2014, 2015, 2018), ICRA (2014, 2016), ICCP (2011), and serves as an Associate Editor of TPAMI. His research has been featured by various media outlets including The New York Times, BBC, MSNBC, Popular Science, and in technology media such as WIRED, The Verge, and New Scientist.
\end{IEEEbiography}

\begin{IEEEbiography}[{\includegraphics[width=1in,height=1.25in,clip,keepaspectratio]{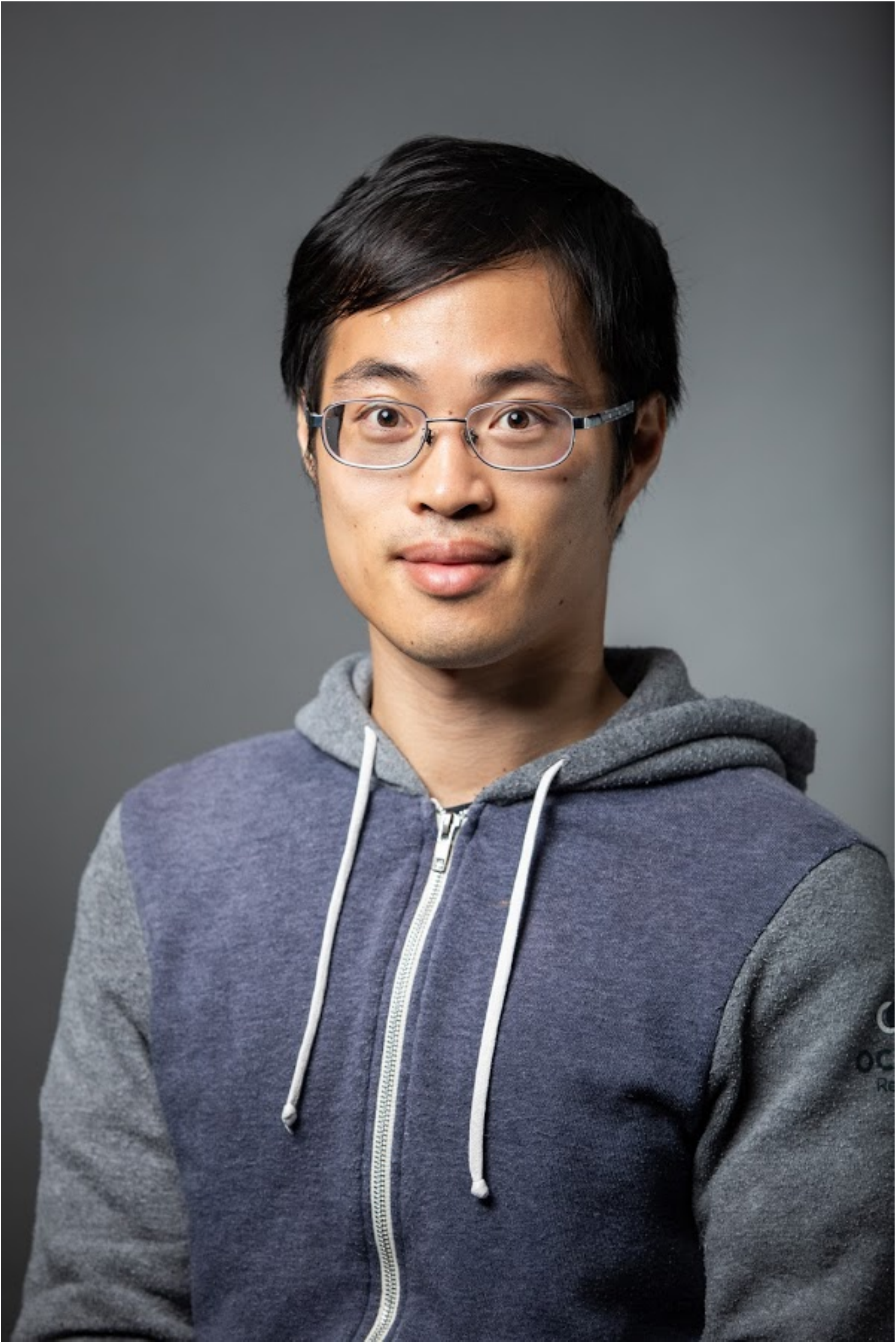}}]{Shoou-I Yu} is currently a research scientist focused on machine learning and computer vision for social VR in Facebook Reality Labs, Pittsburgh, Pennsylvania. Prior to that, he received the Ph.D. in Language Technologies from the School of Computer Science, Carnegie Mellon University, Pittsburgh. He also received the B.S. in Computer Science and Information Engineering from National Taiwan University, Taipei, Taiwan. His research interests mainly lie in landmark detection, multi-object tracking, and multimedia retrieval.
\end{IEEEbiography}

\end{document}